\documentclass[journal]{IEEEtran}
\usepackage{amsmath,amsfonts}
\usepackage{algorithm2e}
\usepackage{array}
\usepackage{textcomp}
\usepackage{stfloats}
\PassOptionsToPackage{hyphens}{url}
\usepackage{verbatim}
\usepackage{graphicx}
\usepackage{cite}
\usepackage{hyperref}
\usepackage{url}
\usepackage{booktabs}
\usepackage{tabularx}
\usepackage{braket} 
\usepackage[dvipsnames]{xcolor}
\usepackage{flushend}
\usepackage{xcolor}
\usepackage{soul}
\usepackage{subcaption}
\usepackage{hyperref} 
\SetKwComment{Comment}{/* }{ */}
\RestyleAlgo{ruled}

\usepackage{xcolor} 

\usepackage{multicol}
\usepackage{multirow}

\usepackage{mwe}
\usepackage{float}
\usepackage{balance}
\usepackage{tikz}
\usetikzlibrary{quantikz}
\usepackage{xfrac}
\usepackage{amssymb}

\usepackage{tikz,xcolor,hyperref}

\definecolor{lime}{HTML}{A6CE39}
\DeclareRobustCommand{\orcidicon}{
	\begin{tikzpicture}
	\draw[lime, fill=lime] (0,0) 
	circle [radius=0.16] 
	node[white] {{\fontfamily{qag}\selectfont \tiny ID}};
	\draw[white, fill=white] (-0.0625,0.095) 
	circle [radius=0.007];
	\end{tikzpicture}
	\hspace{-2mm}
}
\foreach \x in {A, ..., Z}{
	\expandafter\xdef\csname orcid\x\endcsname{\noexpand\href{https://orcid.org/\csname orcidauthor\x\endcsname}{\noexpand\orcidicon}}
}

\protect

\newcommand{\orcid}[1]{\href{https://orcid.org/#1}{\includegraphics[width=4pt]{orcid.png}}}

\begin{document}
\title{
AquaCubeAI-Powered Monitoring\\ Turbidity on-board $\Phi$sat-2 }


\author{\IEEEauthorblockN{
Pietro Di Stasio \orcidA{}, 
Francesca Razzano\orcidB{}, 
Elisa Liparulo\orcidC{}, \IEEEmembership{Student Member, IEEE},\\
Gabriele Meoni\orcidD{}, 
Nicolas Longépé\orcidE{}, 
Deodato Tapete\orcidF{}, \IEEEmembership{Member, IEEE},\\
Paolo Gamba\orcidH{}, \IEEEmembership{Fellow, IEEE},
Gilda Schirinzi\orcidG{}, 
Silvia Liberata Ullo\orcidI{}, \IEEEmembership{Senior Member, IEEE}
}

\thanks{Pietro Di Stasio, Elisa Liparulo, Silvia Liberata Ullo are with the Department of Engineering, University of Sannio, 82100 Benevento, Italy (email:   p.distasio@studenti.unisannio.it; e.liparulo@studenti.unisannio.it; ullo@unisannio.it);\\
Francesca Razzano, Gilda Schirinzi are with the Department of Engineering, University of Naples 'Parthenope', 80133 Naples, Italy (email: francesca.razzano002@studenti.uniparthenope.it; gilda.schirinzi@uniparthenope.it);
\\Gabriele Meoni is with the \(\Phi\)-lab and with the Advanced Concepts and Studies Office, European Space Agency, 00044, Frascati, Italy (email: gabriele.meoni@esa.int);
\\Nicolas Longépé is with the \(\Phi\)-lab, European Space Agency, 00044, Frascati, Italy (email: nicolas.longepe@esa.int);
\\  Deodato Tapete is with the Italian Space Agency (ASI), Via del Politecnico, 00133 Roma, Italy (email: deodato.tapete@asi.it)
\\  Paolo Gamba is with the Department of Electrical, Computer and Biomedical Engineering, University of Pavia, 27100 Pavia, Italy (email: paolo.gamba@unipv.it)}}


\date{}
\maketitle

\begin{abstract}
Timely monitoring of coastal water quality is critical for environmental protection, yet conventional satellite workflows rely on downlink and ground processing, introducing latency that can limit responsiveness to rapidly evolving turbidity events. To address this limitation, we propose AquaCubeAI, a lightweight machine-learning approach for onboard estimation of coastal water turbidity from $\Phi$sat-2 multispectral imagery. By shifting inference from the ground segment to the satellite, AquaCubeAI aims to enable lower-latency, more responsive, and more operationally useful turbidity monitoring under the strict compute and bandwidth constraints of spaceborne platforms. 
The model is trained on simulated $\Phi$sat-2 acquisitions spatially aligned with Copernicus Marine Service (CMEMS)  High-Resolution Ocean-Color (HR-OC) turbidity products over selected localized coastal sites spanning four European marine macro-regions. To provide a realistic evaluation of generalization in the presence of spatial correlation, we adopt a spatial block splitting protocol that mitigates data leakage between training and evaluation subsets. The main contributions of this work are: (i) a scalable dataset generation pipeline pairing simulated $\Phi$sat-2 multispectral patches with CMEMS HR-OC turbidity labels across selected localized European coastal sites; (ii) a compact Multi-Layer Perceptron (MLP)-based turbidity regressor trained under a leakage-aware geospatial split and tailored to embedded constraints; and (iii) a reformulation for dense spatial prediction via parameter sharing, enabling turbidity mapping and simple threshold-based anomaly masks for onboard decision logic. Embedded deployment on an Intel Myriad Vision Processing Unit (VPU) further confirms the feasibility of low-power hardware and supports low-latency inference from multispectral inputs.
Overall, the results obtained on simulated $\Phi$sat-2 acquisitions demonstrate the potential of AquaCubeAI to support efficient and scalable low-latency coastal turbidity monitoring in intelligent Earth Observation missions, while validation on real in-orbit $\Phi$sat-2 observations remains future work.
\textcolor{red}{Data and code will be opened upon acceptance of the paper. Limited access can be granted during the review process upon request.}
\end{abstract}


\section{Introduction}

\IEEEPARstart{W}{ater} is one of the most essential yet increasingly endangered natural resources on the planet, with its quality directly affecting ecosystems, food security, and public health~\cite{6894913, CHARTZOULAKIS201588}. The United Nations has identified universal access to clean water and sanitation as a core objective of the 2030 Agenda for Sustainable Development (SDG~6), yet nearly half of the world's population still experiences severe water scarcity for at least one month each year. International bodies such as the World Health Organization (WHO)~\cite{pruss2008safer} and the United Nations Environment Programme (UNEP)~\cite{unep2021water} underscore the severe health burden associated with inadequate water quality and call for enhanced, large-scale monitoring capacity. These pressures are compounded by climate change, which amplifies hydrological variability and the frequency of extreme events that accelerate water degradation~\cite{10892190, 10941134}.

Despite these priorities, many coastal and marine regions still lack adequate monitoring infrastructure~\cite{Yuan2025, SALAH2025}. Traditional \textit{in situ} techniques, based on manual sampling and laboratory analysis, remain the accuracy benchmark, but are labor-intensive and spatially constrained, making them unsuitable for real-time, large-scale assessments~\cite{cai2022using}. Point-based methods fail to capture the rapid spatial and temporal variability characteristic of coastal environments, highlighting the need for more scalable and responsive monitoring strategies.

Remote Sensing (RS) technologies, especially satellite-based observations, address this gap by providing continuous, synoptic coverage of key optical water quality parameters such as turbidity, chlorophyll-\textit{a}, and colored dissolved organic matter (CDOM)~\cite{yang2023improving}. When integrated with Artificial Intelligence (AI), particularly with Machine Learning (ML) approaches, RS enables the data-driven estimation of these parameters from spectral reflectance, overcoming the limitations of traditional empirical or semi-analytical bio-optical models that struggle to generalize across diverse aquatic environments and sensor characteristics~\cite{ijerph15091881, s20072125}. In particular, AI-based models can capture complex nonlinear spectral–concentration relationships and support automated anomaly detection for near real-time early warning systems~\cite{razzano2024monitoring, mauro2025quantum, liparulo2025impact}.
Despite this progress, most RS-based AI applications rely on ground-based processing chains, which introduce significant latency. This delay primarily stems from the inability to download large amounts of data without direct visibility of a ground station, forcing operations to wait until the satellite approaches the next available contact point. 
Once the data reach the ground, limited prioritization and finite processing capacity within the ground segment may create substantial bottlenecks, particularly when large volumes of EO data must be handled under time-critical operational constraints \cite{barretta2026realtime}, \cite{9288809}.
Moreover, thematic processing frameworks, such as those implemented by CMEMS, tend to rely on bulk processing approaches (e.g., daily and global processing), which further restrict responsiveness and flexibility in managing rapidly evolving contamination events.
A growing body of research addresses these constraints by deploying AI directly onboard satellites for edge processing. This approach enables the selective downlink of only relevant information, such as autonomously detected anomalies, rather than massive volumes of raw data~\cite{di2022early, 9288809, thangavel2022near}. Consequently, onboard processing not only reduces latency but also offers the potential to enable continuous monitoring for missions where coastal and oceanic areas are conventionally excluded from the nominal acquisition plan.
Onboard AI has demonstrated effectiveness across a range of domains \cite{marin2021phi}, including volcanic eruption detection~\cite{del2021board}, vessel detection~\cite{del2023first}, cloud detection~\cite{giuffrida2020cloudscout}, and fire and smoke detection~\cite{lu2024onboard}. However, few works have addressed onboard approaches for marine water quality monitoring~\cite{goudemant2024onboard, maciel2020evaluating}. \\
In this work, the aim is to propose an efficient system able to shift inference from the ground segment to the satellite, enabling lower-latency, and  responsiveness, for timely water quality monitoring. The proposed system builds on a regression-based onboard approach for coastal water quality assessment, as described in previous works by the same authors \cite{liparulo2025impact}, \cite{razzano2024ai}. The new contribution is a lightweight AI-powered system, that we named \textbf{AquaCubeAI}, that estimates turbidity from simulated $\Phi$sat-2 multispectral imagery and generates spatially resolved anomaly masks in an onboard-processing framework.
Under the strict compute
and bandwidth constraints of spaceborne platforms, \textbf{AquaCubeAI} presents the advantage to transmit only the most relevant data to ground stations, supporting the creation of specific alerts. \\
The main contributions of this work are:
\\
(i) a scalable dataset generation pipeline that pairs simulated $\Phi$sat-2 multispectral patches with Copernicus Marine Service (CMEMS) High-Resolution Ocean-Color (HR-OC) turbidity reference targets across selected localized European coastal sites;\\ 
(ii) a lightweight turbidity regressor based on a compact multilayer perceptron (MLP), trained under a leakage-aware geospatial split, suitable for onboard deployment; \\ 
(iii) a reformulation of the regressor for dense spatial prediction through parameter sharing, enabling turbidity mapping and the generation of threshold-based anomaly masks for onboard decision logic;  \\
(iv) an embedded deployment on an Intel Myriad 2 Vision Processing Unit (VPU), the AI accelerator utilized onboard the $\Phi$sat-2 mission, demonstrating low-latency inference from multispectral inputs and confirming feasibility on low-power space hardware with flight heritage.\\
The remainder of this paper is organized as follows. Section~\ref{sec:related_works} reviews related work on water-quality monitoring, while Section~\ref{sec:datasets} describes the dataset generation and preprocessing pipeline based on the CMEMS products. Section~\ref{sec:methodology} details the proposed AquaCubeAI framework, including the training strategy and the reformulation for spatial prediction. Experimental results and embedded inference on Myriad 2 are presented in Section~\ref{sec:results}, and final remarks are provided in Section~\ref{sec:disc_conc}.

\section{Related works on Water Quality Applications with RS and AI techniques} \label{sec:related_works}

Low-latency observation is increasingly recognized as a key requirement in Earth Observation systems designed for dynamic marine environments, where rapid changes may affect ecosystems, coastal infrastructure, fisheries, and human activities within short time windows. In this context, delays introduced by conventional downlink-and-ground-processing pipelines can limit the operational value of satellite monitoring, especially for time-sensitive events such as contaminant dispersion, harmful algal blooms, marine anomalies, and vessel traffic monitoring. These considerations have recently motivated a broader shift toward onboard intelligence, where AI models process data directly in orbit and transmit only actionable information, thus reducing latency and bandwidth demands \cite{barretta2026realtime}.
Traditional in situ monitoring remains the reference standard for water quality assessment because it provides direct and accurate measurements of parameters such as pH, dissolved oxygen, temperature, and nutrient concentrations. Recent developments based on IoT-enabled sensing platforms and autonomous surface vehicles have improved temporal sampling and automation \cite{w13131729}. 

However, these systems still suffer from limited spatial coverage, dependence on field operations, and high maintenance costs, which reduce their scalability in large or remote coastal and marine environments.
RS addresses these limitations by providing synoptic and repeated observations over wide areas. 
Beyond the direct retrieval of water-quality parameters, recent
studies have demonstrated the value of integrating EO data with GIS,
cloud-computing platforms, and GeoAI to develop scalable environmental
monitoring and decision-support tools. Viani et al.~\cite{viani2024one}
developed a Google Earth Engine Web-GIS application based on
Sentinel-2 imagery for the spatial and temporal assessment of water
quality, including turbidity, chlorophyll, and total suspended solids.
In broader environmental applications, Orusa et
al.~\cite{land15040533} combined Copernicus Sentinel-1 and
Sentinel-2 data with a GeoAI framework to support predictive land-cover
mapping and territorial planning. Similarly, the CerMapp platform
proposed by Orusa et al.~\cite{ijgi14110453} demonstrated how
cloud-based geospatial infrastructures can organize field observations
and integrate them with remote-sensing information for spatial
analysis and decision support. Although these applications operate in
different environmental domains, they highlight the increasing role
of interoperable EO, GIS, and AI frameworks in transforming satellite
and geospatial data into operational information.
As reviewed by Mohseni et al.~\cite{MOHSENI2022105701}, satellite and airborne data have become increasingly important for retrieving optical water quality indicators such as chlorophyll-\textit{a}, turbidity, and CDOM. At the same time, Sagan et al.~\cite{SAGAN2020103187} highlighted that RS-based monitoring still faces important limitations, including sensitivity to atmospheric effects, water surface conditions, sensor characteristics, and the need for calibration with \textit{in situ} observations. These challenges have encouraged the integration of AI and ML techniques, which can better capture nonlinear relationships between spectral observations and water quality variables and therefore improve predictive performance in complex aquatic environments.
Within this direction, several studies have demonstrated the potential of AI-enhanced RS for water quality estimation, including the retrieval of chlorophyll-\textit{a}, suspended solids, turbidity, and related bio-optical variables from multispectral or hyperspectral data. These methods have proven particularly useful when frequent \textit{in situ} campaigns are not feasible, but they generally rely on ground-based processing chains and therefore do not fully address latency constraints. In other words, although RS with AI has substantially improved water quality mapping accuracy and spatial coverage, most existing approaches remain oriented toward offline or near-offline analysis rather than true onboard decision support.

Only a limited number of recent works have started to investigate onboard AI for marine and water-related applications. Parsons and Seto~\cite{parsons2024onboard} proposed a Convolutional Neural Network (CNN) for onboard near-real-time harmful algal bloom detection from multispectral satellite imagery, showing the relevance of embedded inference for timely marine hazard identification. Goudemant et al.~\cite{goudemant2024onboard} introduced an onboard anomaly-detection framework for marine environmental protection, targeting potentially unknown pollution events through AI-based onboard analysis. Then, Razzano et al.~\cite{razzano2024ai} explored AI techniques for near-real-time monitoring of coastal contaminants onboard future \(\Phi\)sat-2 acquisitions, demonstrating the feasibility of lightweight models for this application domain. These studies collectively confirm the promise of onboard intelligence for marine monitoring.
A broader perspective also emerges from other latency-sensitive maritime applications, where onboard AI has been investigated for floating debris detection~\cite{ghasemi2023feasibility}, oil-spill identification~\cite{diana2021oil}, ship detection from SAR or optical imagery~\cite{yang2022algorithm, xu2021board, ieracitano2024explainable}, and rapid delivery of maritime information products~\cite{gunzel2025satellite}. Related mission concepts such as EO-Alert~\cite{kerr2021eo} and onboard environmental mapping approaches for water-body recognition~\cite{bui2023swrnet} further support the idea that edge processing can improve responsiveness for marine and maritime domains. Nevertheless, these contributions address detection or classification problems and do not specifically solve the challenge of estimating coastal water quality indicators onboard under realistic RS constraints.


Against this background, the main gap in the literature concerns the lack of onboard frameworks for direct regression of coastal water quality parameters from multispectral satellite observations. Our work addresses this gap by proposing AquaCubeAI, a lightweight onboard framework for turbidity estimation from $\Phi$sat-2 imagery. A first step in this direction was our previous work on onboard regression for water quality monitoring from Razzano et al.~\cite{razzano2024ai}, which laid the foundation for this approach by demonstrating the feasibility of near-real-time onboard estimation using simulated $\Phi$sat-2 data and local in situ measurements from the Ligurian coast. Building on that initial effort, we now present significant improvements in both model architecture and deployment strategy, tackling new constraints that arise in the real-world implementation of the updated model. In particular, the present work differs from the previous one by introducing a scalable dataset generation pipeline based on CMEMS HR-OC turbidity products, a multi-region pilot setting based on selected localized European coastal sites, a leakage-aware spatial block splitting strategy, a more compact embedded regressor with a reduced number of trainable parameters, and a dense spatial prediction formulation for turbidity mapping and threshold-based anomaly masks. Unlike previous works centered on anomaly detection, event recognition, or local proof-of-concept studies, our approach targets a physically meaningful water quality variable, supports spatial prediction through a compact embedded model, and is evaluated through a multi-region pilot study that introduces geographic diversity beyond the previous single-site configuration,
without claiming full representativeness of the corresponding marine
domains, and onboard deployment constraints.

\section{Dataset}
\label{sec:datasets}

The dataset and its generation pipeline were designed to support the proposed onboard turbidity monitoring framework through a scalable pairing of multispectral observations and consistent bio-geochemical reference information. The focus is on coastal and near-shore waters, where turbidity exhibits strong spatial heterogeneity and seasonal variability.

Reference water-quality information is derived exclusively from open-access products distributed by the CMEMS \cite{le2019observation}. Relying on a single, standardized source (instead of local monitoring networks or dedicated \textit{in situ} campaigns) ensures reproducibility and facilitates extension to additional geographical areas.

The input data consist of multispectral $\Phi$sat-2-like imagery generated with the official mission simulator \cite{10623617}, while turbidity labels are obtained from CMEMS HR-OC products. This pairing enables consistent spatial/temporal alignment between satellite observations and reference turbidity, while preserving $\Phi$sat-2 sensor characteristics (e.g., spatial sampling and spectral configuration). 
To introduce geographic variability beyond a single-site case study,
localized Areas of Interest (AoIs) were selected within four European
marine macro-regions (Mediterranean Sea, Black Sea, Iberian coastal region, and North-West Shelf). These sites, sampled across different months to account for seasonal variability, constitute a multi-region pilot dataset and are not intended to represent the complete range of optical water types or environmental conditions occurring throughout each marine domain.
Finally, a dedicated preprocessing pipeline harmonizes CMEMS and simulated $\Phi$sat-2 data into an AI-ready, patch-based dataset compatible with typical onboard computational constraints. Details on CMEMS products, $\Phi$sat-2 data generation, and preprocessing are reported below.

\subsection{Copernicus Marine Service}
\label{subsec:CMEMS}

\begin{figure*}[!t]
    \centering
    \includegraphics[width=1.0\linewidth]{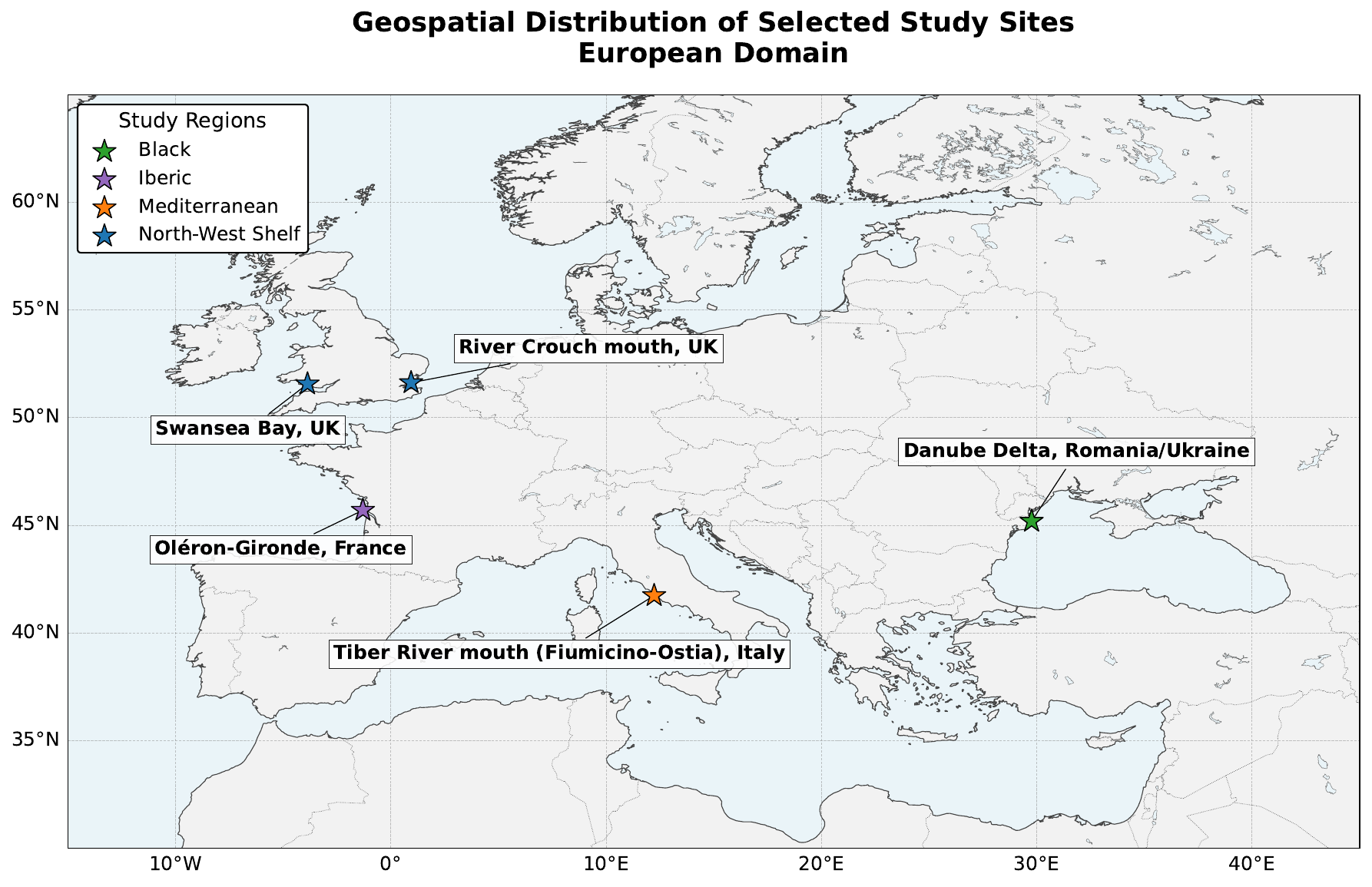}
    \caption{Geospatial distribution of the selected reference locations within the European domain. Markers indicate the localized AoIs included in the pilot study selected within the CMEMS coverage and used to define the corresponding study areas for $\Phi$sat-2 data acquisition.}
    \label{fig:placeholder}
\end{figure*}

Reference turbidity labels were extracted from the CMEMS HR-OC product suite, which provides coastal bio-optical variables derived from Sentinel-2 MSI observations \cite{drusch2012sentinel}. HR-OC products \cite{CMEMS_HROC_PUM} are delivered at $100$~m spatial resolution over a coastal strip extending to approximately 20~km offshore, enabling the observation of fine-scale gradients that are often unresolved by coarser ocean-colour products.
Within HR-OC, turbidity (TUR), expressed in Formazin Nephelometric Units (FNU), represents the primary variable of interest.
TUR belongs to the $tur\_tsm\_chl$ product group together with suspended particulate matter (SPM, in $\text{mg/L}$) and chlorophyll-\textit{a} (CHL, in $\mu\text{g/L}$). These quantities are retrieved from remote sensing reflectance spectra through adaptive switching schemes designed to handle different optical water types \cite{dogliotti2015single}. Since turbidity is sensitive to suspended matter dynamics (e.g., river inputs and resuspension processes), it is a key indicator for coastal water-quality assessment.

In this study, HR-OC turbidity observations acquired during 2023 were extracted from the CMEMS Data Store for four marine regions: the Mediterranean Sea, the Black Sea, the Iberian coastal region, and the North-West Shelf. To balance diversity and dataset size, only localized and representative AoIs were selected within each macro-region. The geographic distribution of the selected AoIs is illustrated in Fig.~\ref{fig:placeholder}, while the reference locations are summarized in Table~\ref{tab:aoi_reference_points}. 
Each reference location was used to define a bounding box for the $\Phi$sat-2 simulation setup, ensuring geographic consistency between HR-OC turbidity labels and the corresponding simulated acquisitions.
The main dataset comprises HR-OC turbidity observations acquired on selected dates between 2 January and 13 November 2023 and retrieved from the CMEMS Data Store. The selected acquisitions span seven months: January, April, June, July, August, October, and November.
In addition, an external hold-out set was constructed from selected acquisition dates between
3 January and 23 December 2024. These samples were excluded from training, validation, and model selection and were used exclusively to assess model generalization under an interannual temporal shift from 2023 to 2024.
For each retained HR-OC observation, the dataset includes acquisition time, geographic coordinates, and the associated turbidity value. These CMEMS-derived measurements are used here as reference turbidity targets for supervised learning. Since they originate from an established remote sensing processing chain rather than from direct \textit{in situ} sampling, part of the residual uncertainty of the CMEMS product may contribute to the overall error budget of the proposed model. At the same time, this choice provides spatially consistent labels at the target resolution and enables reproducible multi-region model development.

\begin{table*}[!ht]
\centering
\caption{Reference locations of the localized coastal sites selected
for the study.}
\label{tab:aoi_reference_points}
\resizebox{0.90\textwidth}{!}{%
\begin{tabular}{llcc}
\toprule
\textbf{Marine macro-region} &
\textbf{Localized study site} &
\textbf{Latitude ($^\circ$)} &
\textbf{Longitude ($^\circ$)} \\
\midrule

Mediterranean Sea &
Tiber River mouth (Fiumicino--Ostia), Italy &
41.7375 & 12.2268 \\

Black Sea &
Danube Delta coastal waters, Romania/Ukraine &
45.1893 & 29.7585 \\

Iberian Coast &
Oléron--Gironde coastal waters, France &
45.7020 & $-1.3000$ \\

North-West Shelf &
River Crouch mouth, United Kingdom &
51.6263 & 0.9372 \\

North-West Shelf &
Swansea Bay, United Kingdom &
51.5718 & $-3.8740$ \\

\bottomrule
\end{tabular}%
}
\end{table*}

\subsection{$\Phi$sat-2 Data Collection}
\label{subsec:phisat2}

Multispectral satellite data consistent with the $\Phi$sat-2 mission were collected to support the proposed onboard turbidity estimation framework. Although $\Phi$sat-2 is currently operational, the dataset used in this study was generated through the official simulation environment made available within the ESA OrbitalAI Challenge \cite{10623617}. This choice was mainly due to the unavailability of the real $\Phi$sat-2 data when this work started, while the simulated data were made available simultaneously with the start of the challenge. 
Although real in-orbit $\Phi$sat-2 observations are now progressively becoming available, a sufficiently consistent set of acquisitions covering the selected Areas of Interest and adequately aligned with suitable turbidity reference targets was not available within the scope of the present multi-region study.
For this reason, the evaluation of the proposed approach on real in-orbit $\Phi$sat-2 observations is left as future work.


$\Phi$sat-2 carries a Visible-to-Near-Infrared (VIS/NIR) multispectral instrument acquiring eight spectral bands (seven multispectral channels and one panchromatic band). 

\begin{table*}[!t]
\centering
\caption{Comparison between the spectral and spatial characteristics
of the $\Phi$sat-2 MultiScape100 payload and the corresponding
Sentinel-2 MSI bands used as input to the official simulator.
Centre wavelength and bandwidth are expressed in nm. The $\Phi$sat-2
ground sampling distance refers to the nominal 500-km orbit.}
\label{tab:phisat2_sentinel2_specs}
\renewcommand{\arraystretch}{1.15}
\setlength{\tabcolsep}{7pt}
\scriptsize
\begin{tabular}{lccclccc}
\toprule
\textbf{$\Phi$sat-2 band} &
\textbf{Centre} &
\textbf{FWHM} &
\textbf{GSD} &
\textbf{Sentinel-2 band} &
\textbf{Centre} &
\textbf{Bandwidth} &
\textbf{Native GSD} \\
&
\textbf{wavelength} &
\textbf{bandwidth} &
\textbf{(m)} &
&
\textbf{wavelength} &
\textbf{(nm)} &
\textbf{(m)} \\
\midrule
MS 1 & 490 & 65  & 4.75 & B2 & 492.4 & 66  & 10 \\
MS 2 & 560 & 35  & 4.75 & B3 & 559.8 & 36  & 10 \\
MS 3 & 665 & 30  & 4.75 & B4 & 664.6 & 31  & 10 \\
MS 4 & 705 & 15  & 4.75 & B5 & 704.1 & 15  & 20 \\
MS 5 & 740 & 15  & 4.75 & B6 & 740.5 & 15  & 20 \\
MS 6 & 783 & 20  & 4.75 & B7 & 782.8 & 20  & 20 \\
MS 7 & 842 & 115 & 4.75 & B8 & 832.8 & 106 & 10 \\
PAN  & 625 & 250 & 4.75 & B2--B6 (synthesized) &  &  &  \\
\bottomrule
\end{tabular}
\vspace{2mm}
\begin{minipage}{0.95\textwidth}
\footnotesize
\textit{\\Temporal characteristics:}
Sentinel-2 has a nominal revisit time of 10 days per satellite and
5 days for the constellation at the Equator. The access frequency of
$\Phi$sat-2 is location- and tasking-dependent, with an average global
access period of approximately 15 days under quasi-nadir conditions.
\end{minipage}
\end{table*}

Table~\ref{tab:phisat2_sentinel2_specs} summarizes the
correspondence between the $\Phi$sat-2 MultiScape100 channels
and the Sentinel-2 MSI bands used as input to the official simulator.
The seven multispectral channels exhibit closely matching centre
wavelengths and bandwidths, which provides the spectral basis for
the simulation workflow. The additional $\Phi$sat-2 panchromatic
band is synthesized through a weighted combination of Sentinel-2
bands B2--B6.

The native Sentinel-2 bands are acquired at spatial resolutions of
10 or 20~m, whereas the simulated $\Phi$sat-2 products have a nominal
ground sampling distance of 4.75~m for all eight channels at the
reference orbital altitude of 500~km. Accordingly, the simulator
performs spatial resampling and introduces sensor-specific effects,
including spectral response adaptation, band-to-band misregistration,
modulation transfer function degradation, and signal-to-noise
variations.

The temporal characteristics of the two missions should be interpreted
separately. Sentinel-2 provides a nominal revisit of 10 days per
satellite and 5 days for the constellation, whereas $\Phi$sat-2 is a
tasked mission whose revisit depends on geographic location and
spacecraft agility, with an average global access period of
approximately 15 days under quasi-nadir conditions. The simulator
reproduces the spectral, radiometric, and spatial characteristics of
the $\Phi$sat-2 payload, but it does not reproduce the actual mission
tasking and revisit constraints; the acquisition dates used in this
study are therefore determined by the availability of the selected
Sentinel-2 observations.

In the official $\Phi$sat-2 simulation chain, Sentinel-2 MSI Level-1C observations are used as input together with scene-classification information and acquisition metadata. The workflow first derives ancillary cloud, cloud-shadow, and cirrus masks, retrieves solar-irradiance and Earth--Sun-distance metadata, and converts Sentinel-2 reflectances into radiance quantities. A $\Phi$sat-2-compatible panchromatic band is then synthesized from the multispectral channels. The simulated imagery is subsequently adapted to the $\Phi$sat-2 payload through spatial resampling, band-to-band misregistration modeling, signal-to-noise degradation, and modulation transfer function filtering. Finally, radiometric processing is applied to generate $\Phi$sat-2-compatible Level-1C reflectance products. This workflow allows the simulator to approximate key spectral, radiometric, spatial, and sensor-related characteristics of the $\Phi$sat-2 payload while preserving the acquisition content of the source Sentinel-2 observations.

\begin{figure*}[!ht]
    \centering
    \includegraphics[width=\linewidth]{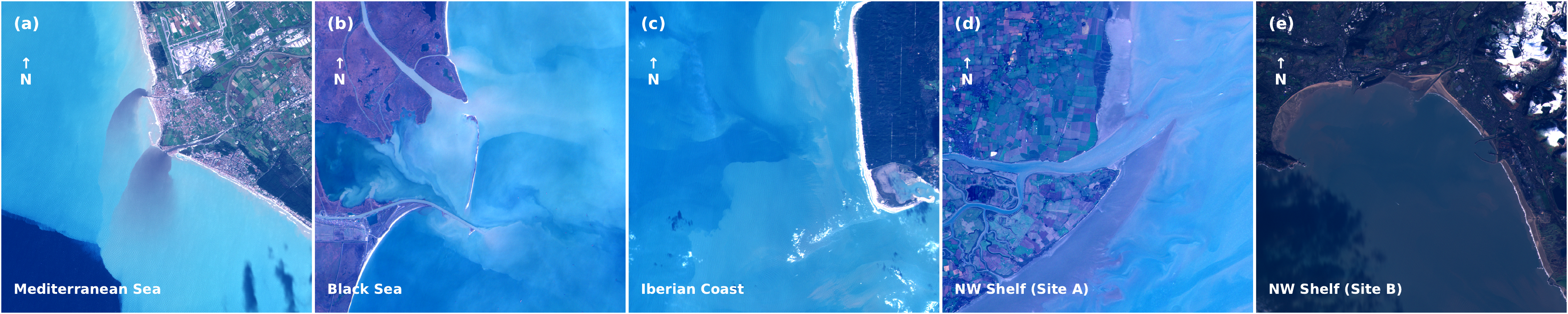}
    \caption{Representative simulated $\Phi$sat-2 true-colour scenes for the selected study areas, illustrating the diversity of coastal environments and optical conditions included in the dataset.}
    \label{fig:phisat-images}
\end{figure*}

Simulated $\Phi$sat-2 Level-1C products were retrieved over the selected AoIs to ensure spatial and temporal consistency with HR-OC turbidity observations. The use of Level-1C data without atmospheric correction is consistent with the intended onboard processing scenario, where computationally intensive atmospheric correction procedures are typically not feasible \cite{furano2020towards}. This choice is not motivated by operational convenience alone. In our recent comparative study, we directly assessed the effect of atmospheric correction on AI-based turbidity estimation and found that Level-1C inputs achieved performance comparable to atmospherically corrected products in the considered coastal setting, with only limited impact on the final prediction accuracy \cite{liparulo2025impact}. In addition, our previous work further confirmed the effectiveness of AI-based turbidity estimation workflows based on Sentinel-2 data for water-quality monitoring applications \cite{razzano2024monitoring}. Taken together, these results support the adoption of Level-1C reflectance in the present work as a practical trade-off between onboard feasibility and predictive performance. Representative simulated $\Phi$sat-2 true-colour scenes are shown in Fig.~\ref{fig:phisat-images}.

\subsection{Dataset Pre-Processing}
\label{subsec:preprocessing}

A dedicated pre-processing pipeline was implemented to harmonize HR-OC turbidity observations with simulated $\Phi$sat-2 multispectral imagery and to construct an AI-ready dataset suitable for onboard regression tasks. The pipeline ensures accurate spatial correspondence between reference turbidity measurements and satellite observations, while preserving physical consistency across data sources and complying with typical onboard computational constraints.


The workflow starts from HR-OC turbidity data. During the $\Phi$sat-2 simulation stage, a scene-level cloud-coverage threshold is applied in the Sentinel Hub acquisition process, and ancillary cloud, cirrus, and cloud-shadow masks are generated from Sentinel-2 scene-classification information. HR-OC observations are then filtered by removing missing values, non-finite values, negative turbidity labels, and turbidity values above the retained range of 140~FNU. For each acquisition date and Area of Interest (AoI), the remaining HR-OC observations are spatially intersected with the footprint of the corresponding simulated $\Phi$sat-2 scene. The simulator generates Level-1C multispectral tiles of $4096\times4096$ pixels; only HR-OC samples falling within the simulated tile coverage and for which a valid patch can be extracted are retained. Although the ancillary masks are preserved as additional layers, they are not used to reject individual patches during dataset preparation.

To achieve precise pixel-level alignment, explicit handling of coordinate reference systems is performed. HR-OC observations are provided in geographic coordinates (WGS84), whereas simulated $\Phi$sat-2 tiles are expressed in a projected coordinate system. When required, HR-OC coordinates are transformed into the image reference system and mapped to pixel indices of the $\Phi$sat-2 tile.

Once a $\Phi$sat-2 tile is available for a given AoI and date, a tiling operation is performed to extract local patches centred on the georeferenced HR-OC observations. For each retained turbidity sample, a $20\times20$ multispectral patch is extracted from the $4096\times4096$ tile at the projected HR-OC location. Given the nominal ground sampling distance of $4.75$~m, each patch covers approximately $100\times100$~m, closely matching the nominal $100$~m resolution of HR-OC products \cite{bailey2006multi}. Each patch includes all spectral channels and is uniquely associated with its turbidity label. No dedicated water/non-water segmentation or preliminary scene classification is introduced at this stage. Patch selection is driven exclusively by the retained valid HR-OC turbidity observations; therefore, the extracted samples are naturally restricted to locations where CMEMS provides valid water turbidity values.

Overall, this pre-processing pipeline integrates heterogeneous data sources into a unified dataset representation, providing a consistent basis for supervised learning and subsequent deployment of turbidity estimation models in onboard processing scenarios.

\subsection{Dataset Analysis and Statistical Characterization}
\label{sec:dataset_analysis}

The constructed dataset was analyzed to quantify its statistical properties and to verify representativeness across heterogeneous coastal conditions. This step is crucial to assess whether the resulting samples provide a reliable basis for training and evaluating onboard regression models, where generalization across space and season is a primary requirement.

\begin{figure}[!ht]
    \centering
    \includegraphics[width=0.98\linewidth]{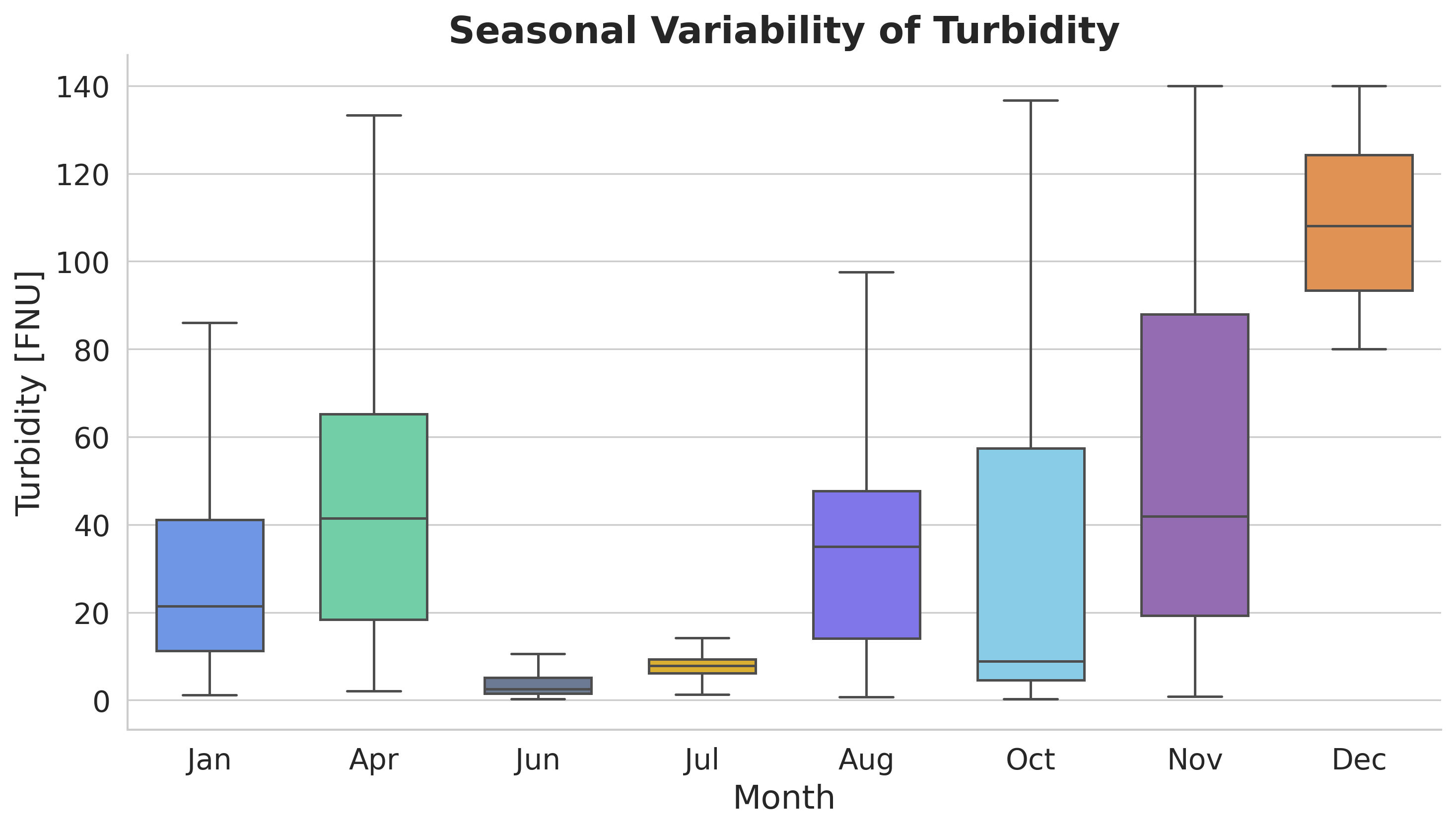}
    \caption{Monthly turbidity variability (FNU), showing seasonal shifts and extremes.}
    \label{fig:seasonal_variability}
\end{figure}

\begin{figure}[!ht]
    \centering
    \includegraphics[width=0.98\linewidth]{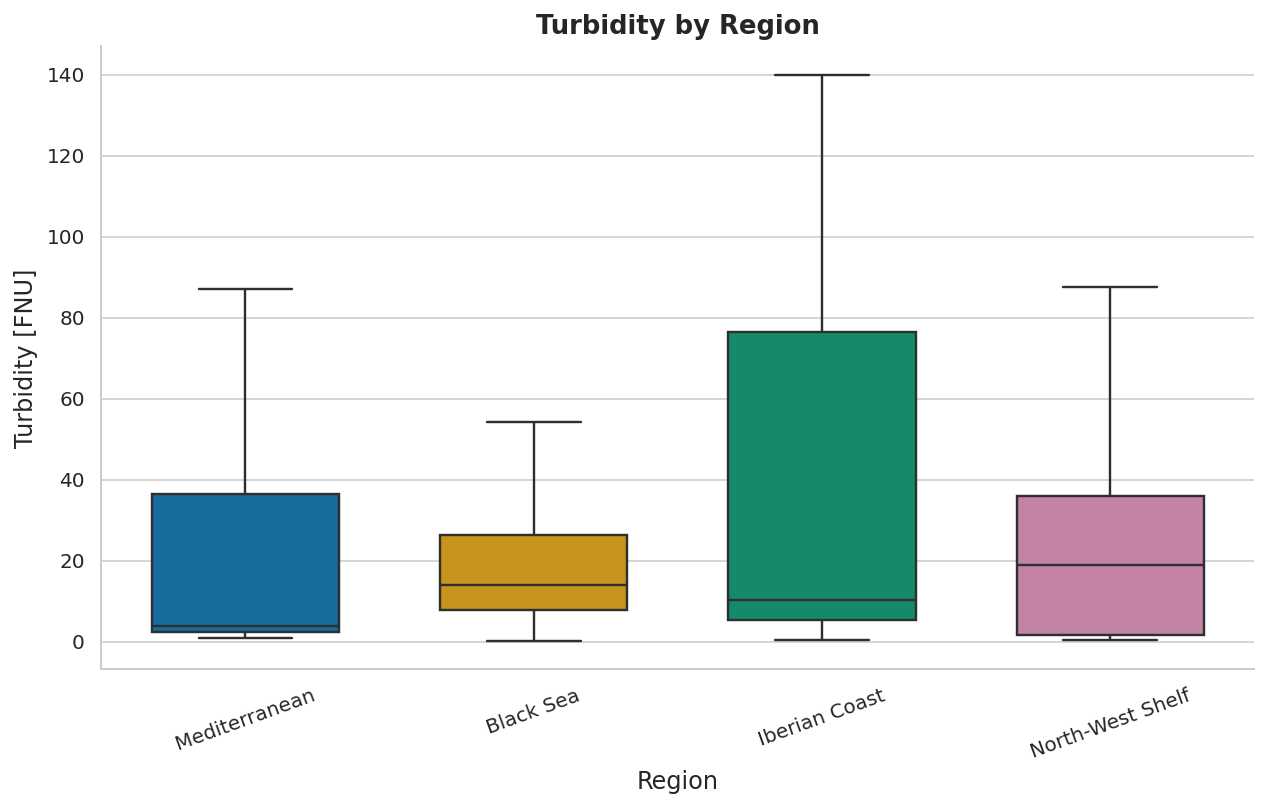}
    \caption{Regional turbidity levels (FNU) across the selected study areas.}
    \label{fig:turbidity_by_region}
\end{figure}

Overall, turbidity values are mostly concentrated in the low-to-moderate range, while high-turbidity observations occur less frequently. This behaviour is consistent with coastal dynamics, where extreme events are less common and often associated with episodic processes, such as river discharge or sediment resuspension under wind-wave forcing \cite{fettweis2012weather}. Nevertheless, the dataset covers a broad turbidity range, including both nominal conditions and high-turbidity regimes of operational relevance.
For clarity, throughout this study, low-to-moderate turbidity refers
to the dataset-specific interval
$0\leq\mathrm{TUR}\leq100$~FNU, whereas high turbidity or the upper
tail refers to $100<\mathrm{TUR}\leq140$~FNU. Within the
low-to-moderate range, the $0$--$40$~FNU interval contains the largest
number of observations and therefore represents the most densely
sampled operating regime. These intervals are specific to the present
dataset and should not be interpreted as universal water-quality
classes.
Seasonal variability is explicitly captured by design. Fig.~\ref{fig:seasonal_variability} summarizes the distribution of turbidity values across the selected months, highlighting marked differences in both central tendency and spread. Spatial heterogeneity is also reflected in the dataset composition. Fig.~\ref{fig:turbidity_by_region} reports turbidity distributions aggregated by region, confirming distinct regimes and reducing the risk of geographic over-specialization \cite{pahlevan2020seamless}.

Finally, the adopted representation preserves the mean spectral information over the local support of each reference observation while providing a compact input format compatible with onboard constraints.

\section{Methodology}
\label{sec:methodology}

This section describes the end-to-end methodology adopted to train and deploy an onboard-capable turbidity regression model using simulated $\Phi$sat-2 multispectral imagery and CMEMS turbidity products as reference. The design is guided by three requirements: \emph{(i)} consistent spatial support between satellite inputs and CMEMS reference measurements; \emph{(ii)} realistic generalization assessment by mitigating spatial leakage \cite{meyer2022machine}; \emph{(iii)} low computational complexity for edge/onboard inference \cite{mateo2021towards}.

Building upon our previous baseline study~\cite{razzano2024ai}, this work extends the learning framework to a CMEMS-driven multi-region pilot dataset composed of selected localized coastal sites and introduces a robust geospatial split strategy. In addition, an embedded-oriented workflow is adopted to produce turbidity estimates and derived spatial products (e.g., anomaly masks) under typical onboard constraints.

\begin{figure*}[!ht]
    \centering
    \includegraphics[width=0.5\linewidth]{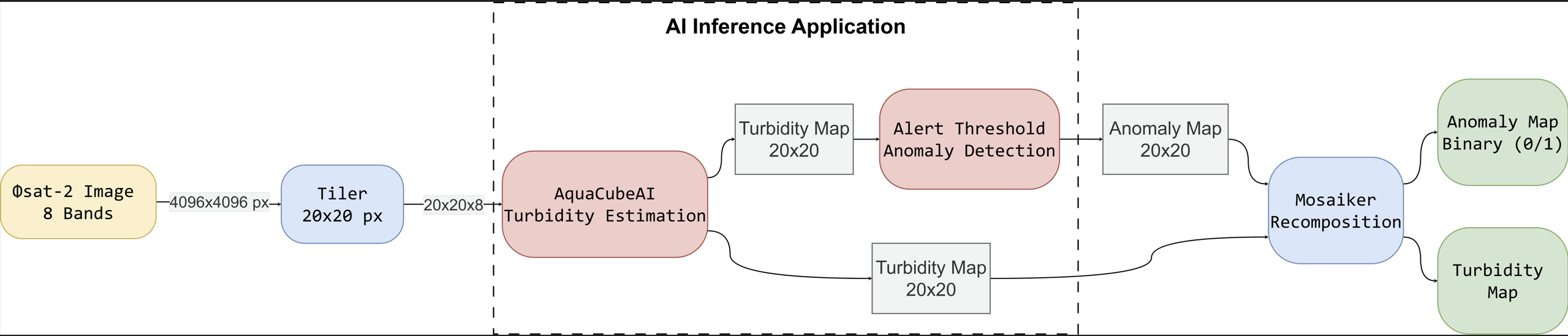}
    \caption{Overview of the proposed onboard turbidity mapping workflow.}
    \label{fig:method_overview}
\end{figure*}

Fig.~\ref{fig:method_overview} summarizes the proposed pipeline: large $\Phi$sat-2 tiles are partitioned into $20\times20$ patches aligned with CMEMS turbidity samples, a lightweight regressor is trained on compact patch descriptors under a geospatially robust split, and the learned model is then reformulated for spatial prediction to generate patch-wise turbidity maps and threshold-based anomaly masks that can be mosaicked into larger-area products and prioritized for downlink.

\subsection{Data Representation}
\label{sec:dataRepresentation}

Each CMEMS turbidity observation is paired with a co-located simulated $\Phi$sat-2 multispectral patch, ensuring consistent spatial support between inputs and labels (Section~\ref{sec:datasets}). The task is formulated as a regression problem, where turbidity is estimated from the multispectral content of the patch.

To keep the representation compact and robust to small geolocation uncertainties, each patch is summarized by a channel-wise spatial average:
\begin{equation}
\mathbf{z}=\frac{1}{HW}\sum_{i=1}^{H}\sum_{j=1}^{W}\mathbf{X}_{i,j,:},
\end{equation}
where $\mathbf{X}\in\mathbb{R}^{H\times W\times C}$ denotes the multispectral patch and $C$ is the number of spectral channels. In this study, C=8, corresponding to the eight $\Phi$sat-2 bands, namely seven multispectral bands and one panchromatic band. Therefore the
panchromatic band was included. The resulting descriptor $\mathbf{z}\in\mathbb{R}^{C}$ captures the mean spectral signature over the CMEMS-aligned area.

Targets are normalized using min--max scaling computed exclusively on the training subset. The scaling parameters are stored and reused for validation/test evaluation and for converting predictions back to physical units (FNU) during inference.

\subsection{Training Strategy}
\label{sec:trainingStrategy}

To obtain reliable performance estimates across geographically heterogeneous coastal environments, the training protocol mitigates spatial leakage and ensures stable optimization over a wide turbidity range. Since remote sensing samples are spatially correlated, purely random splits can lead to optimistic estimates due to spatial autocorrelation \cite{meyer2018improving}. Therefore, a group-aware geospatial splitting strategy (spatial block splitting) is adopted \cite{roberts2017cross}.

\begin{figure*}[!ht]
    \centering
    \includegraphics[width=0.9\linewidth]{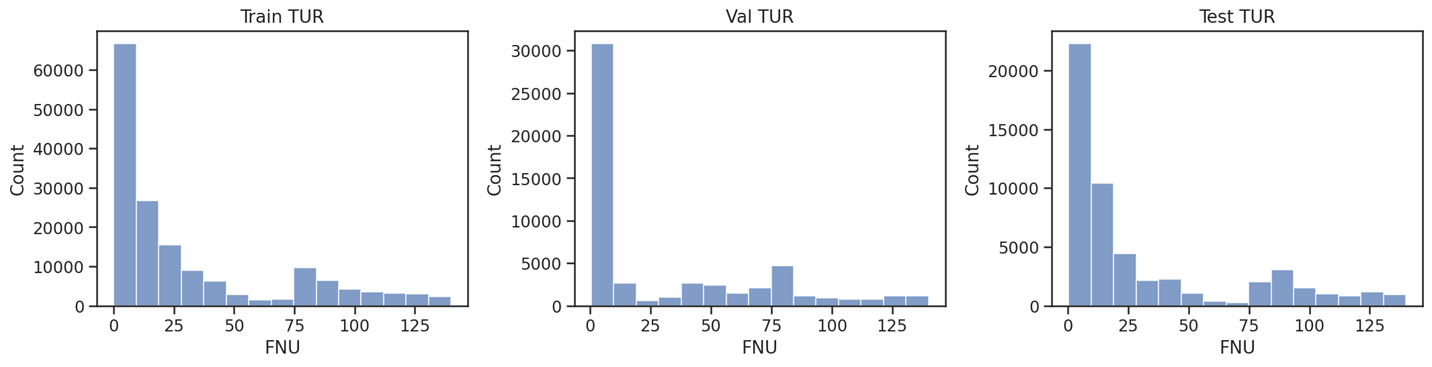}
    \caption{Turbidity distributions (FNU) in the train/validation/test subsets obtained with the proposed geospatial split. The similar distributional profiles across the three subsets confirm that the splitting strategy preserved the overall turbidity balance while maintaining spatial separation.}
    \label{fig:tur_hist_splits}
\end{figure*}

Each sample is assigned to a spatial group by discretizing its UTM coordinates into a regular grid with a 15~km cell size. The 15~km cell size was selected as a pragmatic compromise: it is large enough to reduce short-range spatial dependence between neighboring samples, while still preserving a sufficient number of spatial groups for a stable and balanced partition across the selected study sites. All samples within the same grid cell are treated as a single unit and assigned entirely to either training, validation, or test. No additional temporal constraint was imposed to force all samples from the same acquisition date into a single subset. Therefore, the adopted protocol mitigates spatial leakage, but it does not fully eliminate potential temporal leakage. The final partition is obtained through a greedy randomized assignment with multiple restarts, targeting a $60\%/20\%/20\%$ split while minimizing deviations in split sizes and preserving global turbidity statistics (in particular the mean) across subsets. Fixed seeding is used throughout to ensure reproducibility. Fig.~\ref{fig:tur_hist_splits} reports the turbidity distributions for the three subsets.

\subsection{Regression Model and Optimization}
\label{sec:regressionModel}

The regression function $f_\theta(\cdot)$ is implemented as a lightweight multilayer perceptron (MLP) mapping $\mathbf{z}\in\mathbb{R}^{C}$ to a scalar turbidity estimate. Several configurations are explored to balance predictive capability and computational cost; the default architecture employs three hidden layers with widths $(512,512,256)$ and ReLU activations (Table~\ref{tab:mlp_arch}).

Training is performed with the Adam optimizer and early stopping driven by validation performance. To improve robustness to occasional label noise and rare extreme events, the model is optimized using the Huber loss:
\begin{equation}
\mathcal{L}_\delta =
\begin{cases}
\frac{1}{2}(y - \hat{y})^2, & |y - \hat{y}| \le \delta \\
\delta \left(|y - \hat{y}| - \frac{1}{2}\delta \right), & \text{otherwise},
\end{cases}
\end{equation}
with $\delta=1.0$ in normalized space. This value was selected in normalized space so that the Huber loss retains a smooth MSE-like behavior for typical residuals, while reducing sensitivity to occasional large errors. Gradient clipping and deterministic seeding are adopted to improve stability and guarantee repeatable experiments. The main training hyperparameters are summarized in Table~\ref{tab:train_setup}.

\begin{table}[!ht]
\centering
\caption{Training setup and hyperparameters for the proposed turbidity regressor.}
\label{tab:train_setup}
\resizebox{0.8\columnwidth}{!}{%
\begin{tabular}{lc}
\toprule
\textbf{Parameter} & \textbf{Value} \\
\midrule
Target normalization & Min--max\\
Optimizer & Adam \\
Learning rate & $1\times10^{-4}$ \\
Batch size & 2048 \\
Max epochs & 2000 \\
Early stopping patience & 200 \\
Loss function & Huber ($\delta=1.0$) \\
Gradient clipping & 0.5 \\
Random seed & 5 \\
\bottomrule
\end{tabular}%
}
\end{table}

\begin{table}[!ht]
\centering
\caption{Default MLP architecture used for turbidity regression.}
\label{tab:mlp_arch}
\resizebox{0.6\columnwidth}{!}{%
\begin{tabular}{lc}
\toprule
\textbf{Parameter} & \textbf{Value} \\
\midrule
Input features & $\mathbf{z}\in\mathbb{R}^{C}$ \\
Hidden layers & (512, 512, 256) \\
Activation & ReLU \\
Output & Scalar turbidity $\hat{y}$ \\
\bottomrule
\end{tabular}%
}
\end{table}

\subsection{Model Selection}
\label{sec:modelSelection}

To balance predictive performance and onboard constraints, multiple MLP configurations were tested by varying the number of hidden layers and the size of each layer, while keeping the training protocol and the robust geospatial split fixed. The final model was chosen based on validation RMSE/MAE and parameter count, favoring compact architectures that preserve accuracy. The quantitative comparison and the selected configuration are reported in Section~\ref{sec:results}.

\subsection{From Regression to Spatial Prediction}
\label{sec:spatialPrediction}

\begin{figure*}[!ht]
    \centering
    \includegraphics[width=\linewidth]{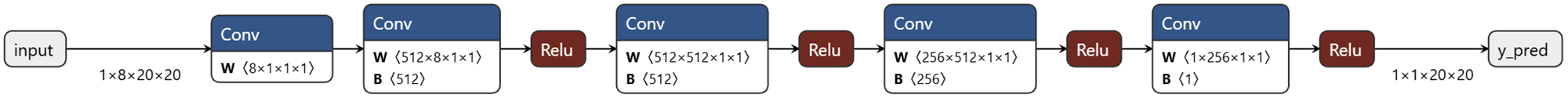}
    \caption{AquaCubeAI model architecture for spatial turbidity prediction.}
    \label{fig:onnx_model_arch}
\end{figure*}

Although the regressor is trained using patch-averaged spectral descriptors, onboard monitoring benefits from spatially resolved outputs, i.e., compact turbidity maps enabling simple downstream logic (e.g., threshold-based flagging). To this end, the trained regressor is reformulated into a spatial predictor based on a CNN, which constitutes the proposed AquaCubeAI model. This architectural choice specifically exploits the hardware acceleration of the VPU. Since the VPU is inherently optimized for CNNs and visual operations, this reformulation allows to effectively parallelize and accelerate the generation of the regression maps. 
Specifically, the trained MLP is reformulated as a sequence of $1\times1$ convolutional layers and applied to the multispectral patch tensor using parameter sharing across spatial locations. Each $1\times1$ layer performs the same spectral transformation at every pixel, producing a dense $H\times W$ turbidity map from an input patch \cite{long2015fully}. While training uses the patch-averaged descriptor, the learned mapping is applied locally via parameter sharing to produce a dense output, without introducing additional trainable parameters or requiring further training.
In our workflow, the AquaCubeAI model is exported to ONNX and compiled for embedded inference (Section~\ref{sec:myriad}). Fig.~\ref{fig:onnx_model_arch} illustrates the resulting ONNX network used to generate turbidity maps; anomaly masks are obtained by thresholding the predicted turbidity.

\section{Results}
\label{sec:results}

This section reports quantitative and qualitative results obtained under the robust geospatial split described in Section~\ref{sec:methodology}. Unless otherwise stated, all metrics are reported in physical units (FNU) after denormalization. We first compare alternative MLP configurations for model selection, then analyze performance across turbidity ranges and against the previous baseline, and finally evaluate robustness on unseen data, spatial mapping products, and embedded deployment on Myriad 2.

\subsection{Architecture Comparison and Selected Model}
\label{subsec:arch_comp_results}

To support model selection, multiple MLP configurations were compared while keeping the training protocol fixed. Among the tested candidates, we also included an exploratory bottleneck configuration inspired by the compressed hidden-layer design adopted in our previous study \cite{razzano2024ai}. Validation results are summarized in Table~\ref{tab:arch_comp}. The $(512,512,256)$ model achieves the lowest validation RMSE among the tested candidates and is therefore retained as the reference configuration for the remainder of this work.

\begin{table}[!ht]
\centering
\caption{MLP architecture comparison on the validation set (metrics in FNU).}
\label{tab:arch_comp}
\setlength{\tabcolsep}{4pt}
\renewcommand{\arraystretch}{1.05}
\resizebox{0.9\columnwidth}{!}{%
\begin{tabular}{@{}lcccc@{}}
\toprule
\textbf{Architecture} & $\mathbf{R^2}$ & \textbf{RMSE} & \textbf{MAE} & \textbf{Bias} \\
\midrule
\textbf{(512, 512, 256)} & \textbf{0.995} & \textbf{2.78} & \textbf{1.41} & $-0.34$ \\
(512, 256, 128) & 0.995 & 2.81 & 1.44 & $-0.45$ \\
(512, 512, 256, 128) & 0.994 & 2.95 & 1.62 & $-0.54$ \\
(512, 512, 512, 512, 512, 512) & 0.993 & 3.19 & 1.67 & $-0.72$ \\
(512, 512, 256, 128, 64) & 0.994 & 2.97 & 1.64 & $-0.86$ \\
(512, 512, 512, 512, 512, 43, 43) & 0.985 & 4.76 & 3.58 & $-0.96$ \\
(128, 128, 128, 128) & 0.989 & 4.06 & 2.22 & $-1.11$ \\
(32, 64, 32) & 0.980 & 5.30 & 3.14 & $-1.19$ \\
(32, 64, 128, 256, 128, 64, 32) & 0.931 & 10.23 & 6.59 & $-3.86$ \\
\bottomrule
\end{tabular}%
}
\end{table}

\begin{figure*}[!t]
\centering
\begin{subfigure}{1.0\linewidth}
\centering
\includegraphics[width=\linewidth]{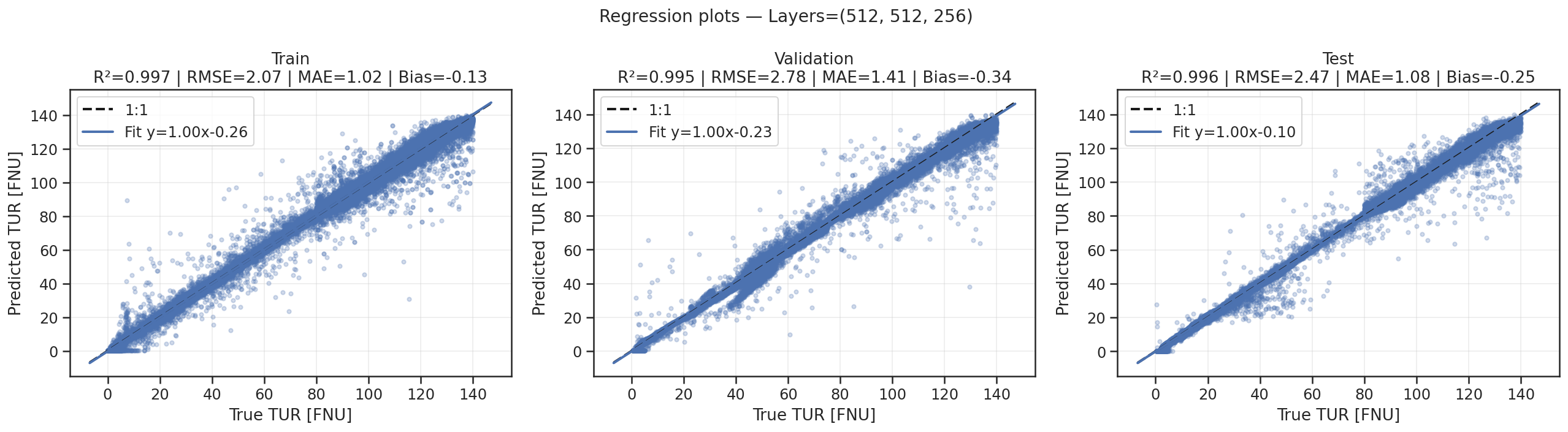}
\end{subfigure}

\vspace{2mm}

\begin{subfigure}{1.0\linewidth}
\centering
\includegraphics[width=\linewidth]{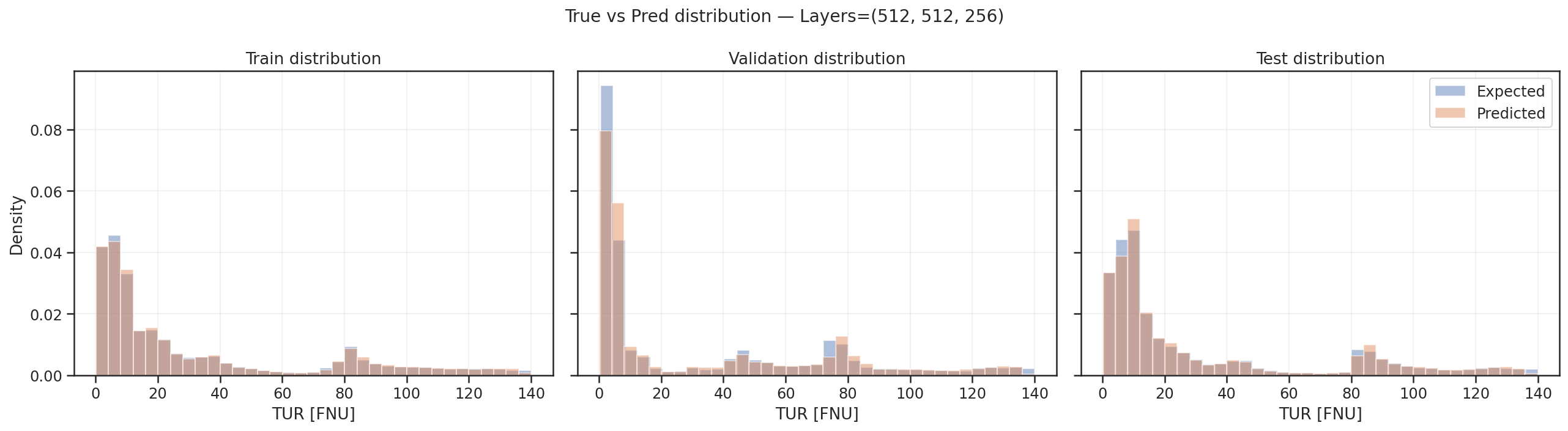}
\end{subfigure}

\caption{Selected model $(512,512,256)$ on the robust train/validation/test split: (top) predicted vs.\ reference turbidity scatter plots, and (bottom) reference vs.\ predicted turbidity distributions.}
\label{fig:results_best}
\end{figure*}

Fig.~\ref{fig:results_best} qualitatively summarizes the behaviour of the selected model on the robust split. Scatter plots (top) show agreement between predicted and reference turbidity over the full range, while the distribution plots (bottom) indicate that the regressor reproduces the overall turbidity statistics, including the skewed low-to-moderate regime and the less frequent high-turbidity tail.

\subsection{Performance Across Turbidity Ranges and Baseline Comparison}
\label{subsec:ranges_baseline}

To characterize model behaviour across operating regimes, Table~\ref{tab:metrics_by_range_val} reports validation metrics computed on subsets obtained by masking samples according to the \emph{reference} turbidity value.

The proposed regressor achieves its best performance in the most represented turbidity range ($0$--$40$~FNU), where RMSE and MAE are lowest. 
However, from an operational perspective, the higher-turbidity regime is particularly relevant, as it is more closely associated with the events of greatest monitoring interest.
As the upper bound of the considered interval increases (e.g., $0$--$80$ and $0$--$100$~FNU), errors increase gradually, reflecting the inclusion of rarer and more heterogeneous high-turbidity conditions.
This behaviour is expected in coastal environments: elevated turbidity levels are typically linked to localized and transient processes (e.g., river plumes, sediment resuspension, nearshore mixing), which introduce stronger spatial variability and increase the difficulty of learning a globally consistent spectral-to-turbidity mapping. In addition, the reduced number of high-turbidity samples limits the model’s ability to fully capture extreme regimes, resulting in a mild degradation of accuracy at the upper tail of the distribution, while preserving stable performance in the operational low-to-moderate regime. Despite the gradual increase in RMSE and MAE as the evaluated range expands, the model maintains strong overall agreement with the reference values across the complete 0–140 FNU interval, achieving an $R^2$ of 0.995. This result indicates that AquaCubeAI preserves a high capability to reproduce the overall variability of turbidity values, although the larger errors observed toward the upper tail confirm that extreme conditions remain more challenging.

\begin{table}[!ht]
\centering
\caption{Performance by turbidity range on the validation set (mask on reference turbidity).}
\label{tab:metrics_by_range_val}
\setlength{\tabcolsep}{4pt}
\renewcommand{\arraystretch}{1.05}
\resizebox{0.8\columnwidth}{!}{%
\begin{tabular}{lrrrrr}
\toprule
\textbf{Range (FNU)} & \textbf{N} & $\mathbf{R^2}$ & \textbf{RMSE} & \textbf{MAE} & \textbf{Bias} \\
\midrule
$0$--$40$  & 35250 & 0.959 & 1.39 & 0.86 & $-0.35$ \\
$0$--$80$  & 47301 & 0.994 & 2.04 & 1.13 & $-0.23$ \\
$0$--$100$ & 50230 & 0.994 & 2.24 & 1.21 & $-0.19$ \\
$0$--$140$ & 54395 & 0.995 & 2.78 & 1.41 & $-0.34$ \\
\bottomrule
\end{tabular}%
}
\end{table}

Compared with our previous work~\cite{razzano2024ai}, which was conducted on a local case study, the present work extends
the evaluation to a multi-region pilot setting based on selected
localized European coastal sites and under a stricter leakage-aware geospatial split strategy. This broadens the geographic scope
of the analysis without implying full representativeness of the
optical variability within the corresponding marine domains.
This shifts the analysis from a localized proof of concept toward a more geographically diverse and methodologically rigorous evaluation scenario. For comparability with the previous study, we additionally report results restricted to the $0$--$40$~FNU range, which was used there as the main quantitative reference.
While the experimental conditions are not fully matched, reporting the same operating range provides a useful reference point. In this regime, the proposed model attains RMSE $=1.39$~FNU and MAE $=0.86$~FNU on the current validation subset (Table~\ref{tab:comp_baseline_0_40}).


Notably, AquaCubeAI achieves these results with a substantially reduced parameter count, decreasing from $818{,}349$ parameters in the baseline to $401{,}417$ parameters here, which benefits onboard deployment in terms of memory footprint and compute.

\begin{table}[!ht]
\centering
\caption{Performance comparison with~\cite{razzano2024ai} in the $0$--$40$~FNU range.}
\label{tab:comp_baseline_0_40}
\setlength{\tabcolsep}{4pt} 
\renewcommand{\arraystretch}{1.05}
\resizebox{\columnwidth}{!}{%
\begin{tabular}{lcccc}
\toprule
\textbf{Method} & \textbf{Parameters} & \textbf{RMSE} & \textbf{MAE} & \textbf{Geographic coverage} \\
\midrule
AI4EDoET~\cite{razzano2024ai} & 818{,}349 & 1.10 & 0.74 & Local (Ligurian coast) \\
AquaCubeAI & 401{,}417 & 1.39 & 0.86 & Multi-region (EU coastal sites) \\
\bottomrule
\end{tabular}%
}
\end{table}




\subsection{External Hold-Out Evaluation}
\label{subsec:unseen}

\begin{figure}[!ht]
    \centering
    \includegraphics[width=0.8\linewidth]{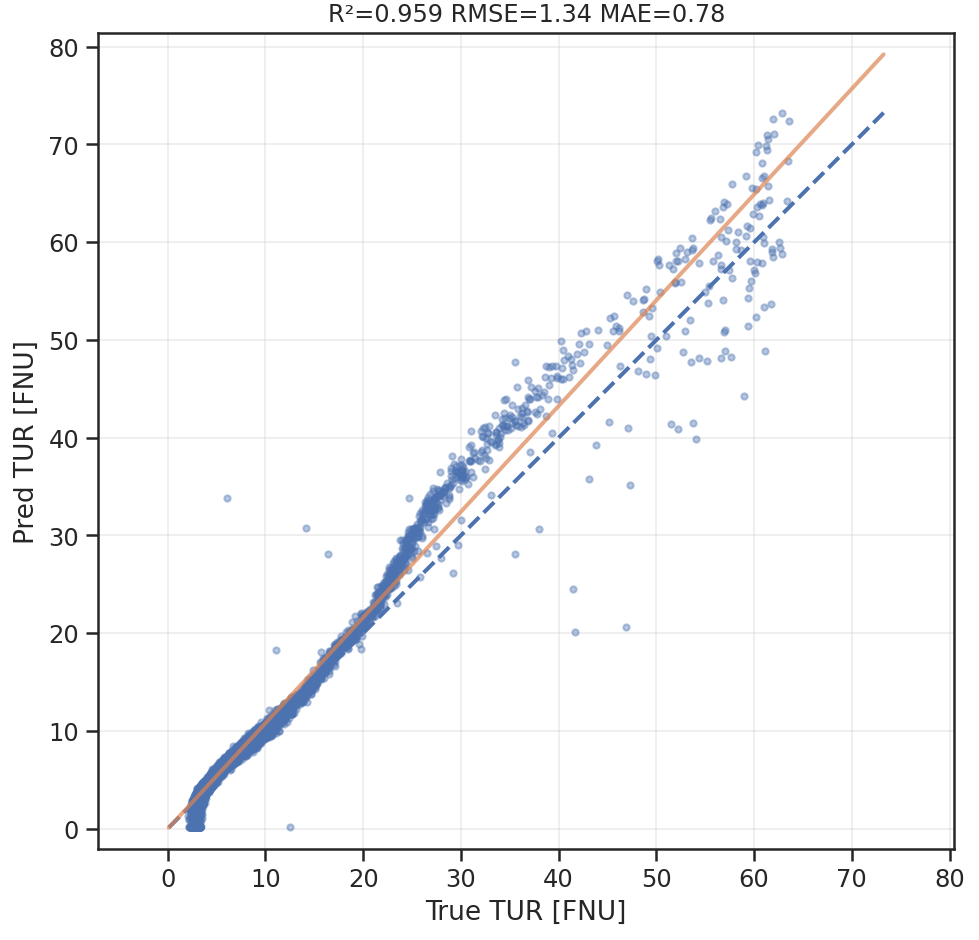}
    \caption{External hold-out evaluation: predicted vs.\ reference turbidity}
    \label{fig:unseen_test}
\end{figure}

To assess robustness under distribution shifts, we evaluated the selected model on an independent acquisition from the North-West Shelf region that was excluded from training, validation, and test. The same preprocessing pipeline was applied. Under these conditions, the external hold-out evaluation provides additional evidence of the model's robustness on independent data. As shown in Fig.~\ref{fig:unseen_test}, predictions remain tightly clustered around the 1:1 line, with the corresponding performance metrics reported directly in the figure. Residual dispersion is mainly observed at higher turbidity levels, where fewer samples are available and local variability is stronger.
To provide an estimate of the variability associated with random
initialization and stochastic training, the complete training procedure
was repeated using five independent random seeds, while keeping the
data split, preprocessing pipeline, architecture, and hyperparameters
fixed. Across the five runs, the test set yielded an RMSE of
$2.70 \pm 0.17$~FNU, an MAE of $1.25 \pm 0.13$~FNU, and a bias of
$-0.55 \pm 0.23$~FNU. For the external 2024 hold-out
set, the model achieved an RMSE of $1.59 \pm 0.15$~FNU, an MAE of
$0.95 \pm 0.14$~FNU, and a bias of $0.19 \pm 0.23$~FNU.
The limited standard deviations indicate that the reported performance
is stable with respect to the training seed. 

\subsection{Global Mapping and Spatial Products}
\label{subsec:global_mapping}
Before evaluating the spatial products generated through the
formulation described in Section~\ref{sec:spatialPrediction}, we
assessed whether retaining within-patch spatial context could improve
patch-level turbidity estimation. The proposed patch-averaged MLP was
therefore compared with a lightweight CNN operating directly on the
full $20\times20\times8$ input patch. The CNN comprises two
$3\times3$ convolutional layers with ReLU activations, followed by
global average pooling and a linear regression head. Both models were
trained and evaluated using the same geospatial split, target
normalization, and evaluation protocol.

The CNN did not improve performance on any of the evaluated splits.
The proposed MLP achieved lower RMSE than the CNN on the validation
set (2.78 versus 3.28~FNU), test set
(2.47 versus 3.51~FNU), and external hold-out set
(1.34 versus 1.43~FNU). The MLP also achieved lower MAE and higher
$R^2$ values across all three splits. These results indicate that,
within the lightweight onboard-constrained setting considered here,
introducing limited local spatial context does not improve prediction
accuracy over the proposed patch-averaged representation. The
patch-averaged MLP was therefore retained because it provides both
higher predictive accuracy and a computationally simpler
representation for onboard inference.

\begin{figure*}[!t]
    \centering
    \includegraphics[width=1.0\linewidth]{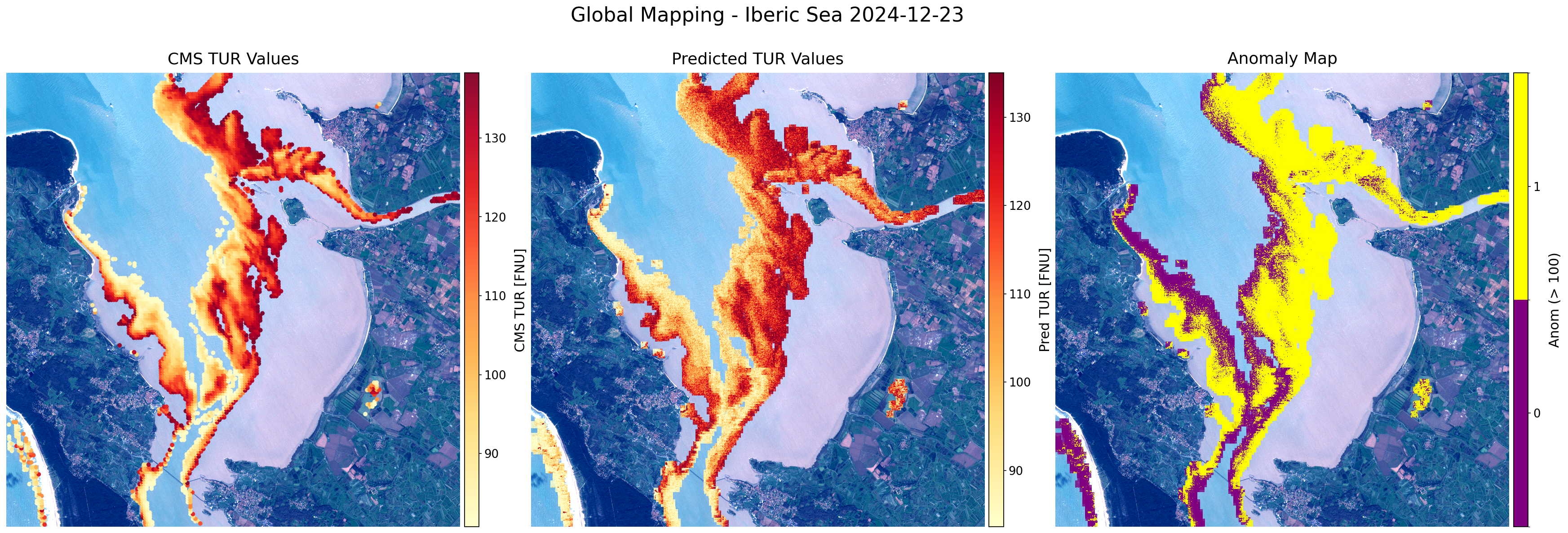}
    \caption{Global mapping example (Iberian AoI, 2024--12--23): CMEMS reference turbidity, AquaCubeAI turbidity map, and the derived anomaly mask ($\mathrm{TUR} > 100$~FNU).}
    \label{fig:global_mapping}
\end{figure*}

We next demonstrate how the spatial formulation described in Section~\ref{sec:spatialPrediction} can be used to produce dense outputs over larger tiles. Fig.~\ref{fig:global_mapping} shows an example over the Iberian AoI (2024--12--23). The first two panels compare the CMEMS reference turbidity with the corresponding model turbidity map, obtained by applying the spatial predictor to each patch and mosaicking the resulting outputs over the tile footprint.

The third panel illustrates a compact decision-oriented product derived from the pixel-wise turbidity map: an anomaly mask generated via thresholding ($\mathrm{TUR} > 100$~FNU \cite{InSituTurbidityValues}). This example highlights the practical value of the proposed onboard workflow, where dense estimates can be turned into lightweight alert layers to support event flagging and data prioritization for downlink.

\begin{figure*}[!b]
    \centering
    \includegraphics[width=1.0\linewidth]{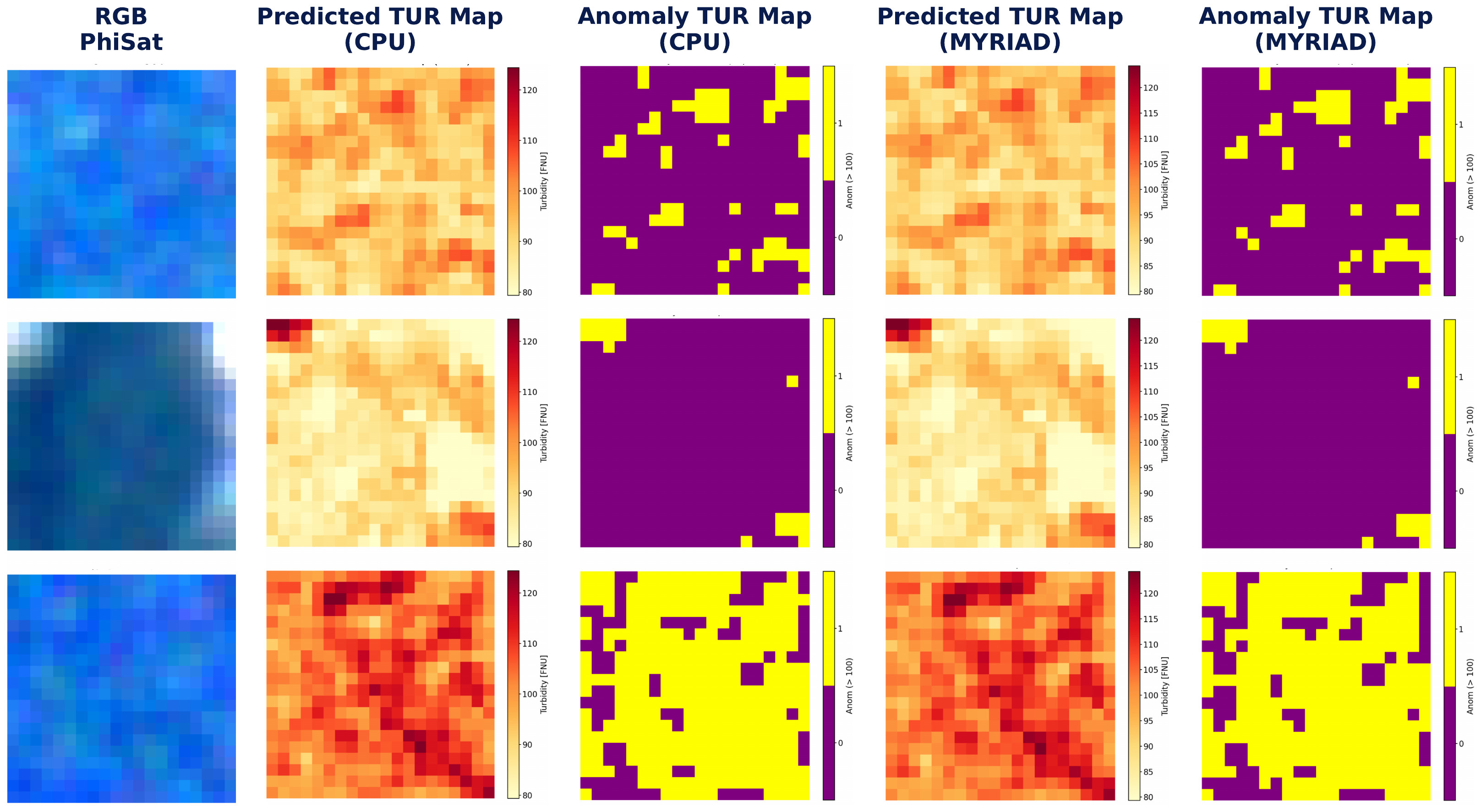}
    \caption{Patch-level inference example: $\Phi$sat-2 RGB patch, turbidity map and anomaly mask on CPU, and turbidity map and anomaly mask on Myriad (anomaly obtained with $\mathrm{TUR}>100$~FNU). The agreement between the CPU and Myriad 2 outputs confirms that the embedded deployment preserves the model predictions.}
    \label{fig:myriad_example}
\end{figure*}
\subsection{Inference on Myriad 2}
\label{sec:myriad}

Embedded inference feasibility was assessed by deploying the spatial model on Myriad~2 through an OpenVINO-based pipeline. We report runtime performance together with the ability of the model to generate the target spatial products, namely turbidity maps and anomaly masks, directly from multispectral inputs. The trained network was exported to ONNX and compiled for the Myriad target. In the current setup, inference is carried out on fixed-size multispectral patches selected according to the valid HR-OC turbidity locations retained during the pre-processing stage, with input tensor shape $[1,C,20,20]$ ($C=8$), producing a dense $20\times20$ turbidity map. A lightweight post-processing step then applies a simple thresholding rule ($\mathrm{TUR}>100$~FNU) to derive a binary anomaly mask for rapid event flagging.
Runtime measurements obtained with our pipeline on Myr\nobreak iad~2 confirm the feasibility of the proposed approach in an embedded setting and indicate low patch-level latency, with the turbidity model achieving, on Myr\nobreak iad~2, an average model-only latency of 8.51~ms per patch, corresponding to 117.49~inferences/s, while processing all retained valid patches extracted from the selected 4096 × 4096 scene required a total of 215.6\nobreak\;s.
For broader onboard applications, this workflow could be further extended with a lightweight preliminary screening stage to automatically identify and discard non-water areas, as suggested by Goudemant et al.~\cite{goudemant2024onboard}. This stage is not part of the current implementation, where patches are selected according to valid HR-OC turbidity observations. This would make water-area identification automatic in a fully onboard setting, further improving efficiency by applying the turbidity model only where needed. Overall, these results confirm the efficiency of the proposed workflow and support its potential applicability in onboard processing scenarios. An illustrative patch-level example, including turbidity maps and anomaly masks generated on CPU and on Myriad 2, is reported in Fig.~\ref{fig:myriad_example}.

\section{Discussion and Conclusions}
\label{sec:disc_conc}

This work introduced \emph{AquaCubeAI}, an onboard turbidity estimation framework trained on simulated $\Phi$sat-2 multispectral data and CMEMS HR-OC turbidity products used as reference. The proposed approach combines a scalable dataset generation pipeline, geographically diverse coastal domains, and a lightweight embedded model designed for near real-time inference under onboard processing constraints. By reformulating the regressor for dense spatial prediction, AquaCubeAI enables the generation of turbidity maps and threshold-based anomaly masks, supporting onboard decision logic and selective data downlink for low-latency water-quality monitoring scenarios.
The patch-averaging strategy was retained as an explicit
accuracy--complexity design choice for onboard inference. Although it
discards sub-patch spatial patterns, the tested lightweight CNN did
not improve prediction accuracy. Moreover, the subsequent
$1\times1$ convolutional reformulation increases the spatial
resolution of the output but does not recover spatial context.

An additional aspect concerns the use of Level-1C inputs together with CMEMS-derived reference turbidity targets. While the adoption of Level-1C reflectance is motivated by onboard constraints and supported by previous comparative analyses, prediction performance may still be affected under atmospheric conditions that are not adequately represented in the training data. In addition, because the supervisory signal originates from the CMEMS HR-OC processing chain rather than from direct \textit{in situ} measurements, residual uncertainties in the reference product may contribute to the overall error budget of the proposed framework.

A further limitation concerns the exclusive use of simulated mission-like $\Phi$sat-2 data throughout training and evaluation. Although the official simulator reproduces key spectral, radiometric, spatial, and sensor-related characteristics of the payload, a residual gap with real in-orbit observations may still arise because of calibration uncertainties, real sensor noise, residual misregistration, atmospheric variability, and mission-operation effects that cannot be fully captured by simulation alone.

Results obtained under the robust geospatial split indicate that accurate turbidity estimation is achievable across heterogeneous coastal environments, with strongest performance in the low-to-moderate operational regime and reduced accuracy in the high-turbidity tail, where fewer samples are available and local variability is stronger. The external hold-out test offers a first evaluation on an unseen acquisition date and suggests that the learned mapping remains informative under temporally unseen conditions. However, since the main geospatial split was not explicitly constrained by date, residual temporal dependence across the internal subsets cannot be fully excluded. Additional validation is therefore required to better characterize robustness under distribution shifts. Embedded tests on an Intel Myriad VPU confirm the feasibility of executing the model on low-power hardware with latencies compatible with near real-time inference.

Future work will extend validation to additional regions and seasons, evaluate performance on real in-orbit $\Phi$sat-2 observations, and improve robustness to rare high-turbidity extremes and distribution shifts. Further efforts will also address dataset expansion and sampling strategies to better represent high-turbidity regimes.

\section*{Acknowledgments}
\noindent This work has been partially supported by the Italian Space Agency (ASI) through the PhD scholarship of Pietro Di Stasio under the contract N. 2024-24-HH.0 CUP: F83C24000890005.

\bibliographystyle{IEEEtran}
\bibliography{ref}
\vspace{-1.0cm}
\begin{IEEEbiography}[{\includegraphics[width=1.32in,height=1.18in,clip,keepaspectratio]{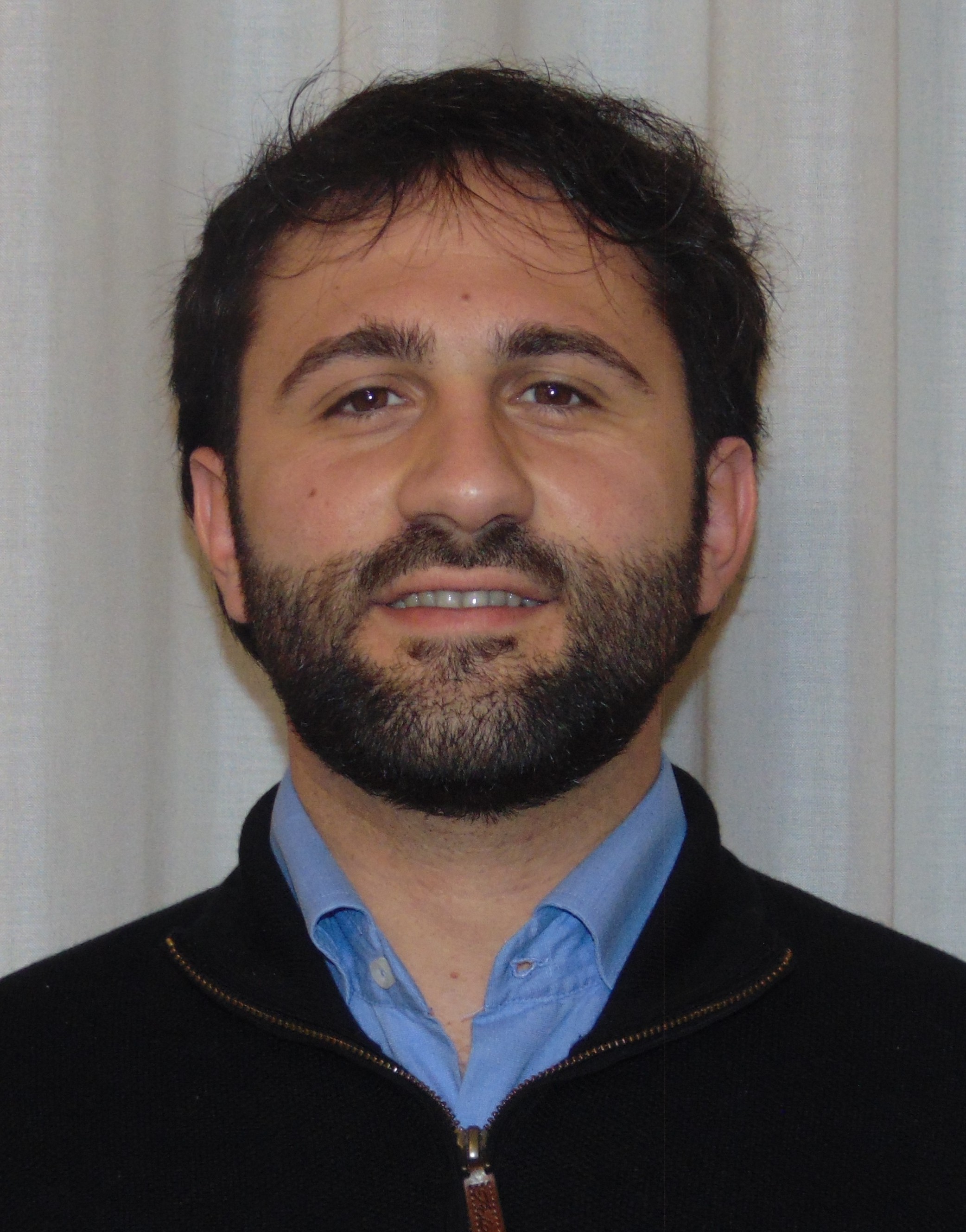}}]{Pietro Di Stasio} is a PhD student in Information Technologies for Engineering at the University of Sannio , supported by a scholarship fully funded by the Italian Space Agency (ASI). He obtained his Master's Degree cum Laude in Electronic Engineering for Automation and Telecommunications from the University of Sannio in September 2024.
His research interests primarily revolve around Earth Observation, Remote Sensing , and the development of lightweight, AI-based on-board processing systems for near real-time response to cascade risks affecting anthropic areas. Following his Bachelor's degree , he was awarded a junior research fellowship in 2022 at the University of Sannio , under the supervision of Prof. Silvia L. Ullo. During this time, he focused on utilizing AI on-board techniques for the early detection of volcanic eruptions and spent part of this period at ESA ESRIN in Frascati, Italy.
Pietro is an active IEEE Student Member with affiliations to the Geoscience and Remote Sensing Society (GRSS) and the Aerospace and Electronic Systems Society (AESS). Additionally, he was a member of a team selected in the Top-5 of the European Space Agency's (ESA) $\Phi Sat-2$ OrbitalAI Challenge.
\end{IEEEbiography}
\begin{IEEEbiography}
[{\includegraphics[width=1in,height=5.55in,clip,keepaspectratio]{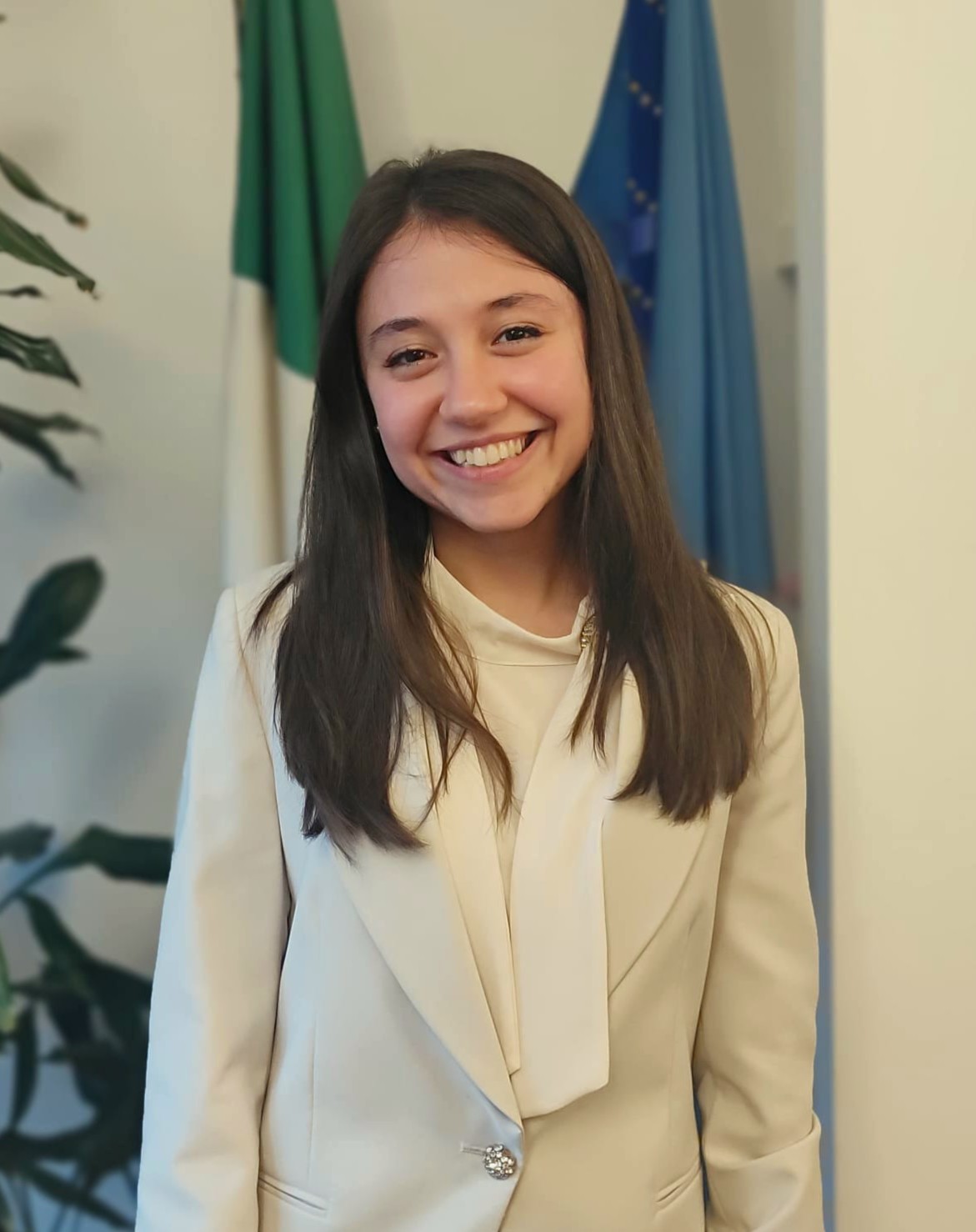}}]{Francesca Razzano} graduated cum laude
in Electronic Engineering for Automation and
Telecommunications from the University of
Sannio in 2023. She is currently enrolled in
the Ph.D. program in Information and Communication Technology and Engineering at the
University of Parthenope, in Naples, under the
supervision of Prof. Gilda Schirinzi and cosupervision of Prof. Silvia L. Ullo. Her research primarily focuses on Remote Sensing
and satellite data analysis, as well as the application of Artificial Intelligence techniques for Earth observation.
In particular, she investigates water quality monitoring and forest
tree height estimation. She has also contributed to research on the
fusion of optical and SAR data for different tasks. In addition,
she works on the development of onboard AI for Remote Sensing
systems. Her professional experience includes a position as a Visiting
Researcher at the European Space Agency’s $\Phi$-Lab. She has coauthored papers and articles presented at renowned conferences in
the field of remote sensing. She is an IEEE Student Member, actively
involved in IEEE GRSS IDEA initiatives, and participates in the
IEEE Young Professionals Affinity Group of the Italy Section as
Treasurer.
\end{IEEEbiography}
\vspace{-1.0cm}

\begin{IEEEbiography}
[{\includegraphics[width=1in,height=1.15in,clip,keepaspectratio]{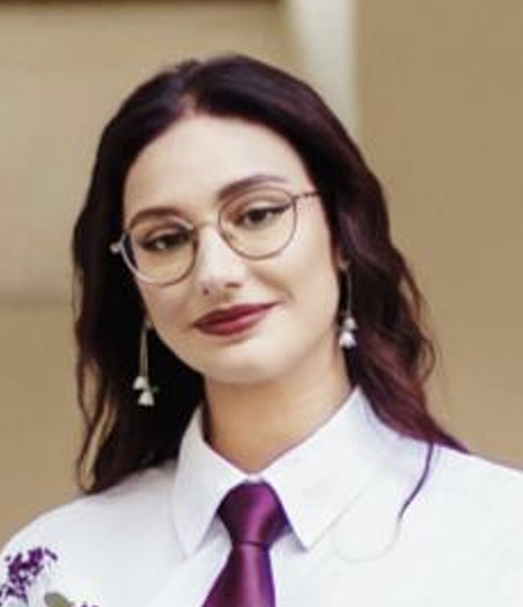}}]{Elisa Liparulo} Master student at University of Sannio, specializing in Computer  Engineering. Her research interests primarily revolve around Earth Observation, Remote Sensing and the application of AI onboard for satellite data analysis.
She was awarded a research scholarship in 2025 at University of Sannio, under Supervisor Prof. Silvia L. Ullo, following her Bachelor's degree.
Her work during this scholarship focused on utilizing AI onboard for water contaminants detection.
\end{IEEEbiography}
\vspace{-1.0cm}

\begin{IEEEbiography}[{\includegraphics[width=1in,height=1.15in,clip,keepaspectratio]{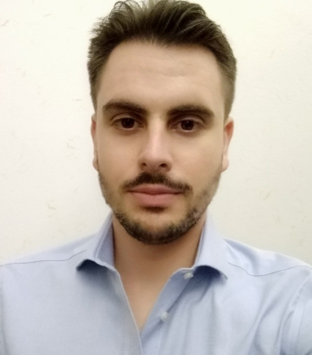}}]{Gabriele Meoni} received his M.Sc. degree
in Electronic Engineering and his Ph.D. degree in Information Engineering from the University of Pisa respectively in 2016 and in
2020, where he was supervised by Prof. Luca
Fanucci. After completing his doctoral studies,
from September 2020 to April 2023 he held
the position of Internal Research Fellow at
the European Space Agency (ESA) (Advanced
Concepts Team (ACT) September 2020 - August 2021, $\Phi$-lab September 2021 - April 2023),
where he conducted research on Artificial Intelligence (AI) and
neuromorphic computing for onboard spacecraft applications. During
2022-2023, he was a visiting researcher at AI Sweden, focusing on
distributed edge learning for satellite constellations. From May 2023
to April 2024, he served as an Assistant Professor in the Faculty of
Aerospace Engineering at Delft University of Technology. Currently,
Meoni is an Innovation Officer at ESA, with research interests
spanning satellite onboard processing, AI for Earth Observation, and
neuromorphic computing. Meoni coauthored more than 40 scientific
publications.
\end{IEEEbiography}
\vspace{-1.0cm}

\begin{IEEEbiography}[{\includegraphics[width=1in,height=1.15in,clip,keepaspectratio]{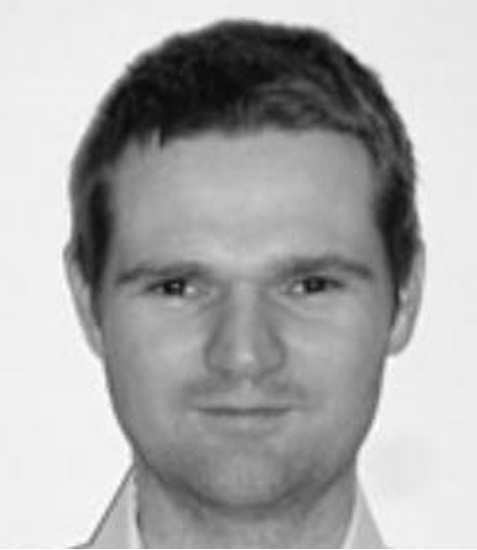}}]{Nicolas Longépé} received his M.Eng. degree
in electronics and communication systems and
M.Sc. degree in electronics from the National
Institute for the Applied Sciences, Rennes,
France, in 2005 and his Ph.D. degree in signal
processing and telecommunication from the
University of Rennes I, Rennes, in 2008. From
2007 to 2010, he was with the Earth Observation Research Center, Japan Aerospace Exploration Agency, Tsukuba, Japan. From 2010
to 2020, he was with the Space Observation
Division, Collecte Localization Satellites, Plouzané, France, where
he was a research engineer. Since September 2020, he has been an
Earth observation Data Scientist,
$\Phi$-Lab Explore Office, European
Space Research Institute, European Space Agency, 00044 Frascati,
Italy. His research interests include Earth observation remote sensing
and digital technologies. such as machine (deep) learning. He has
been working on the development of innovative synthetic aperture
radar-based applications for environmental and natural resource
management (ocean, mangrove, land and forest cover, soil moisture,
snow cover, and permafrost) and maritime security (oil spills, sea ice,
icebergs, and ship detection/tracking). At the
$\Phi$-Lab, he is particularly involved in the development of innovative Earth observation
missions in which artificial intelligence is directly deployed at the
edge (on the spacecraft).
\end{IEEEbiography}
\vspace{-1.0cm}

\begin{IEEEbiography}[{\includegraphics[width=1in,height=1.15in,clip,keepaspectratio]{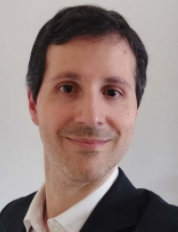}}]{Deodato Tapete} received the Ph.D.
degree in earth sciences, specialized in synthetic aperture radar (SAR) and optical satellite remote sensing
for monitoring of cultural heritage, archaeological
remote sensing, assessment of natural and anthropogenic hazards, urban applications, agriculture and
water resources management, in 2012.
He has been a Researcher in earth observation
and data analytics with the Italian Space Agency
(ASI) since 2017. He has developed several methods
based on SAR data processing to address issues of
heritage conservation, including but not limited to structural stability, impacts
of infrastructure construction, weathering and deterioration, looting, intentional
destruction. He is ASI program scientist for the Committee on Earth Observation
Satellites (CEOS) Working Group on Disaster (WGD). He also leads the ASI
programme “Innovation for Downstream Preparation for Science".
\end{IEEEbiography}
\vspace{-1.0cm}

\begin{IEEEbiography}[{\includegraphics[width=1in,height=1.15in,clip,keepaspectratio]{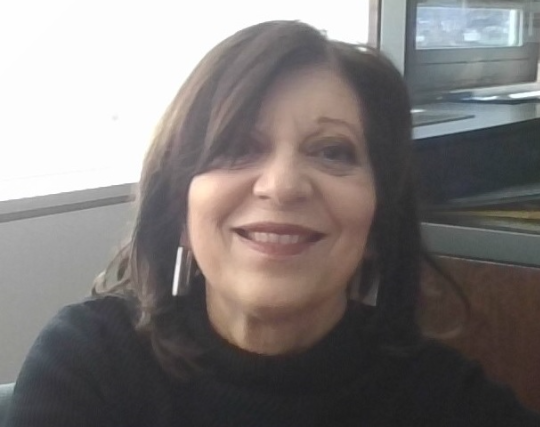}}]{Gilda Schirinzi} graduated cum laude in electronic
engineering at the University of Naples ”Federico
II.” From 1985 to 1986, she was at the European
Space Agency, ESTEC, Noordwijk, The Netherlands. In 1988, she joined the Istituto di Ricerca
per l’Elettromagnetismo e i Componenti Elettronici
(IRECECNR), Naples, Italy. In 1998, she joined
the University of Cassino, Italy, as an associate
professor of telecommunications, and in 2005, she
became a full professor. Since 2008, she has been
at the University of Naples “Parthenope.” Her main
scientific interests are in the field of signal processing for Remote Sensing
applications, with particular reference to synthetic aperture radar (SAR)
interferometry and tomography. She is a Senior Member of the IEEE.
\end{IEEEbiography}
\vspace{-1.0cm}

\begin{IEEEbiography}[{\includegraphics[width=1in,height=1.15in,clip,keepaspectratio]{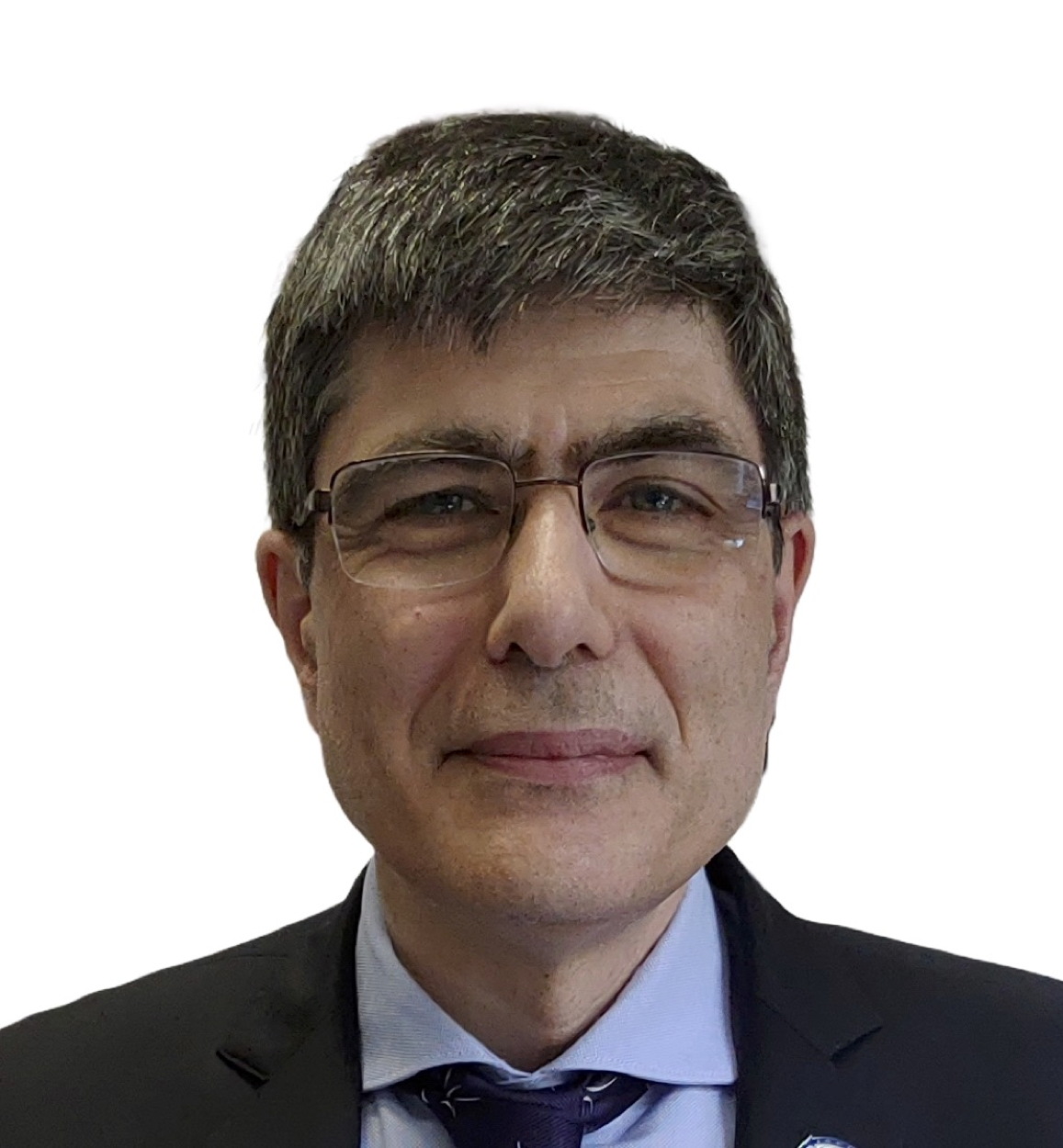}}]{Paolo Gamba} is Professor at the University of Pavia, Italy in the Telecommunications and
Remote Sensing Laboratory. He received the Laurea degree in Electronic Engineering
“cum laude” from the University of Pavia, Italy, in 1989, and the Ph.D. in Electronic
Engineering from the same University in 1993.
He served as Editor-in-Chief of the IEEE Geoscience and Remote Sensing Letters from
2009 to 2013 and of the IEEE Geoscience and Remote Sensing Magazine in 2023-2024.
He was Chair of the Data Fusion Committee of the IEEE Geoscience and Remote
Sensing Society (GRSS) from October 2005 to May 2009. He has been elected in the
GRSS AdCom from 2014 to 2022 and served as GRSS President from 2019 to 2020. He
also served as Technical Co-Chair of the 2010, 2015 and 2020 IGARSS conferences, in
Honolulu (Hawaii), Milan (Italy), and on-line, respectively.
He is Fellow of IEEE, IAPR, AAIA and the Academia Europaea. He has been invited to
give keynote lectures and tutorials on several occasions about urban remote sensing, data
fusion, EO data for physical exposure and risk management. He published more than 210
papers in international peer-review journals.
\end{IEEEbiography}
\vspace{-1.0cm}

\begin{IEEEbiography}[{\includegraphics[width=1in,height=1.15in,clip,keepaspectratio]{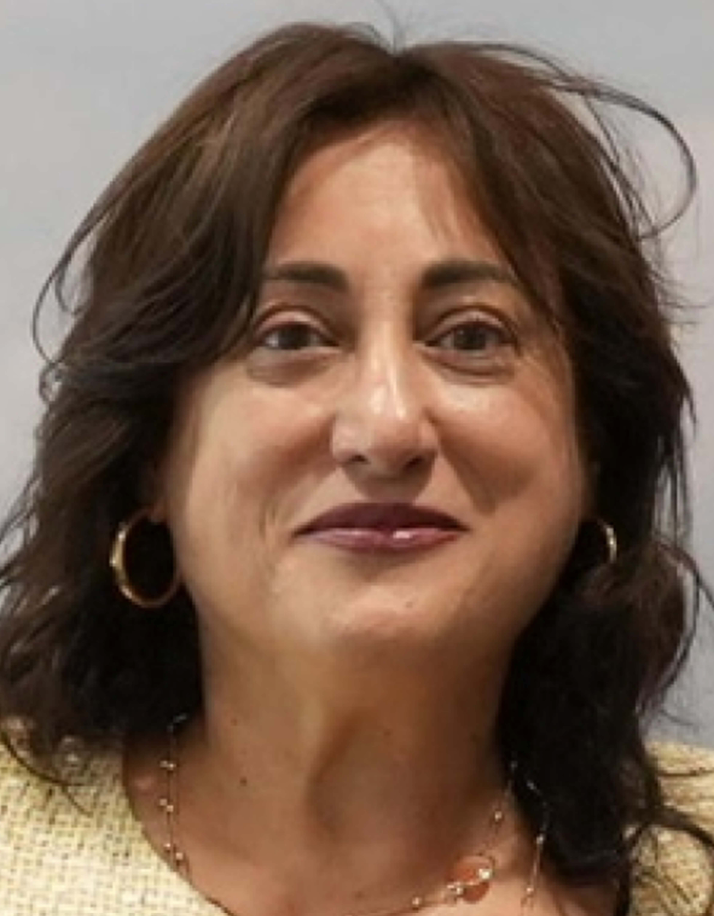}}]{Silvia Liberata Ullo} IEEE Senior Member, IEEE AESS Italy Chapter Chair, IEEE GRSS Europe Liaison and AdCom Member, Member of the Image Analysis and Data Fusion Technical Committee of the IEEE GRSS since 2020 and Chair of the IADF MIA Working Group, Industry Liaison for IEEE Joint ComSoc/VTS Italy Chapter. National Referent for FIDAPA BPW Italy Science and Technology Task Force (2019-2021). 
Graduated with Laude in Electronic Engineering with a specialization in Telecommunications at Federico II University in Naples (Italy), in 1989. She pursued a Master of Science in Management at the Massachusetts Institute of Technology (MIT) in Cambridge (U.S.A.) in 1992. Researcher since 2004 at the University of Sannio, Benevento (Italy), and Associate Professor since 2026. Member of the Academic Senate (until October 2025) and PhD Professors’ Board (present). Courses: Probability and Signals, Geospatial Data Analysis (Bachelor program); Earth monitoring and mission analysis Lab (Master program), Optical and radar Remote Sensing (Ph.D. program).  Authored 140+ research papers, co-authored many book chapters and served as editor of two books. Associate Editor of relevant journals (IEEE TGRS, JSTARS, GRSL, MDPI Remote Sensing, Springer Arabian Journal of Geosciences and others). Co-Editor-in-Chief of IET Image Processing. Guest Editor of many special issues. Research interests: signal processing, radar systems, sensor networks, smart grids, remote sensing, satellite data analysis, machine learning and quantum ML applied to remote sensing.
\end{IEEEbiography}

\end{document}